\documentclass[pdflatex,sn-mathphys-num]{sn-jnl}

\usepackage{graphicx}%
\usepackage{multirow}%
\usepackage{amsmath,amssymb,amsfonts}%
\usepackage{amsthm}%
\usepackage{mathrsfs}%
\usepackage[title]{appendix}%
\usepackage{xcolor}%
\usepackage{textcomp}%
\usepackage{manyfoot}%
\usepackage{booktabs}%
\usepackage{algorithm}%
\usepackage{algorithmicx}%
\usepackage{algpseudocode}%
\usepackage{listings}%

\usepackage{graphicx}%
\usepackage{multirow}%
\usepackage{amsmath,amssymb,amsfonts}%
\usepackage{amsthm}%
\usepackage{mathrsfs}%
\usepackage[title]{appendix}%
\usepackage{xcolor}%
\usepackage{textcomp}%
\usepackage{manyfoot}%
\usepackage{booktabs}%
\usepackage{algorithm}%
\usepackage{algorithmicx}%
\usepackage{algpseudocode}%
\usepackage{listings}%

\usepackage[utf8]{inputenc}
\usepackage[T1]{fontenc}

\usepackage{tikz}%
\usetikzlibrary{positioning, arrows.meta, shapes.misc}%
\usetikzlibrary{mindmap, shadows, backgrounds}%
\usetikzlibrary{shapes}%

\usepackage[most]{tcolorbox}%

\usepackage{microtype}
\usepackage{nicefrac}
\usepackage{geometry}

\usepackage{url}
\usepackage{doi}
\usepackage{hyperref}

\hypersetup{
    colorlinks=true,
    linkcolor=blue,
    citecolor=blue,
    urlcolor=blue,
    pdfborder={0 0 0}
}

\theoremstyle{thmstyleone}%
\theoremstyle{thmstyletwo}%

\theoremstyle{thmstylethree}%

\begin{document}

\title[Article Title]{OpenClaw and Ollama in Agentic AI: Toward Fully Autonomous and Scalable AI Agent Systems}


\author*[1]{\fnm{Konstantinos I.} \sur{Roumeliotis}}\email{k.roumeliotis@uop.gr}

\author*[2]{\fnm{Ranjan} \sur{Sapkota}}\email{rs2672@cornell.edu}

\affil*[1]{\orgdiv{Department of Informatics and Telecommunications}, \orgname{University of the Peloponnese}, \orgaddress{\city{Tripoli}, \postcode{22131}, \country{Greece}}}

\affil*[2]{\orgdiv{Department of Biological and Environmental Engineering}, \orgname{Cornell University}, \orgaddress{\city{Ithaca}, \state{NY}, \postcode{14850}, \country{USA}}}


\abstract{The rapid transition from reactive large language model (LLM) interfaces to persistent, action-capable systems has revealed fundamental gaps in the architectural understanding of Agentic AI, particularly in disentangling inference, orchestration, and execution layers. Despite significant progress, there remains a lack of unified frameworks that systematically explain how autonomous agents can be designed, deployed, and evaluated as full-stack systems. This paper presents a comprehensive, layered analysis of Agentic AI architectures. We begin by establishing the foundations of Agentic AI and layered architectures, examining the evolution from reactive LLMs to persistent agentic systems and identifying core design principles such as modularity, memory hierarchy, and continuous execution. We then analyze OpenClaw and Ollama as a full-stack Agentic AI architecture, where Ollama functions as the LLM inference layer and OpenClaw operates as the agent runtime layer, enabling seamless integration from model inference to autonomous reasoning, planning, and action. To substantiate this architectural perspective, we introduce a prototype experimental validation of the OpenClaw–Ollama full-stack system. Through controlled configurations, benchmark task design, and system-level evaluation metrics, the results demonstrate a consistent monotonic improvement in performance as architectural complexity increases, confirming that autonomous capabilities such as tool use and persistent memory emerge from system integration rather than isolated models. Building on these findings, we further analyze operational challenges, safety, and evaluation, highlighting critical issues in security, privacy, governance, and benchmarking of persistent agentic systems. Finally, we explore future research directions, including scalable multi-agent systems, distributed architectures, and human-centered, responsible autonomous AI. Overall, this work establishes a unified architectural framework for Agentic AI, provides empirical validation of full-stack autonomous systems, and outlines a roadmap toward scalable, secure, and trustworthy autonomous agents. Prototype experimental validation, architecture, models, code, and experimental datasets are openly released to support reproducibility, transparency, and community benchmarking at \href{https://github.com/Applied-AI-Research-Lab/OpenClaw-and-Ollama-in-Agentic-AI}{GitHub}.}


\maketitle

\section{Introduction}
The emergence of large language models (LLMs) has fundamentally reshaped the trajectory of artificial intelligence, shifting the field from narrow task-specific pipelines toward increasingly general-purpose reasoning systems capable of language understanding, code generation, multimodal interpretation, and decision support \cite{chen2024evolution,sarikaya2025path}. Early deployments of LLMs, however, were predominantly reactive: a user issued a prompt, the model generated a response, and the interaction terminated unless a new prompt was manually provided \cite{liu2023prompting}. Such systems demonstrated remarkable linguistic competence, yet they remained limited in persistence, autonomy, memory continuity, and grounded action \cite{zheng2026lifelong}. As a result, a growing gap has become evident between what foundation models can express in natural language and what real-world intelligent systems must do in practice, namely perceive changing environments, maintain long-term objectives, coordinate tools, preserve state across interactions, and act over extended temporal horizons. This gap has catalyzed the rise of \textbf{Agentic AI}, a paradigm in which LLMs serve not merely as text generators but as cognitive engines embedded within broader software architectures for planning, memory management, tool use, and autonomous execution \cite{sapkota2025ai, abou2025agentic}.

Within this transition, the central challenge is no longer only model capability, but system architecture. In other words, the practical realization of autonomous AI agents depends on how inference engines, memory modules, orchestration layers, execution pipelines, communication interfaces, and safety controls are composed into an integrated runtime \cite{alva2026agentic, bandi2025rise}. A single powerful model is insufficient to produce a robust agentic system if it lacks persistent state, reliable access to external tools, governance constraints, or mechanisms for long-horizon reasoning \cite{kumar2026agentic, wei2026agentic}. Consequently, the engineering focus of contemporary AI research is moving toward layered architectures that separate the concerns of model inference, agent orchestration, and environment interaction \cite{alva2026agentic, adabara2025trustworthy}. This architectural shift has become especially important for LLM-based multi-agent systems, in which multiple AI agents may collaborate, specialize, exchange information, and jointly execute complex workflows \cite{li2024survey, he2025llm}. In such settings, modularity, interoperability, and runtime control become essential design goals.

Two emerging technologies illustrate this layered transformation particularly well: \textbf{Ollama} (\href{https://github.com/ollama/ollama}{Source Link}) and \textbf{OpenClaw} (\href{https://github.com/openclaw/openclaw}{Source Link}). Ollama represents the \textbf{LLM inference layer}, providing a lightweight and practical mechanism for running large language models locally, efficiently, and with stronger privacy control \cite{rodriguez2025privacy}. It addresses a major limitation of cloud-dependent LLM deployments by enabling on-device or self-hosted inference, thus improving deployment flexibility, latency characteristics, and data sovereignty \cite{marcondes2025using}. OpenClaw, by contrast, represents the \textbf{agent runtime layer}, offering an architectural framework in which models can be embedded into persistent, tool-using, memory-enabled, and action-capable autonomous agents \cite{weidener2026openclaw, borjigin2026execution}. Its significance lies not in model training, but in its orchestration of cognition, memory, scheduling, and execution into a continuous operational loop \cite{ying2026uncovering, deng2026taming}. When considered together, Ollama and OpenClaw provide a useful lens through which to examine the full-stack realization of \textbf{Agentic AI}: one supplies the computational substrate for inference, while the other organizes that intelligence into persistent agent behavior.

This layered perspective is scientifically important for several reasons. First, it clarifies that modern Autonomous Agents are not monolithic artifacts but composite systems, and that their behavior emerges from interactions among heterogeneous modules rather than from model weights alone \cite{buyya2026agentic,wang2026autonomous}. Second, it enables a more rigorous taxonomy of agentic architectures by distinguishing between the inference layer, the runtime layer, the memory layer, the tool interaction layer, and the governance layer \cite{zhang2026agentic}. Third, it provides a clearer basis for analyzing system-level trade-offs, such as privacy versus scalability \cite{mo2026ironengine, flynn2026cognitive}, local deployment versus cloud elasticity \cite{holzbauer2026malicious, ruan2026logic}, architectural flexibility versus integration complexity, and autonomy versus controllability. Finally, it makes it possible to evaluate systems such as OpenClaw and Ollama not merely as isolated tools, but as representative components of a broader infrastructural stack for Agentic AI, AI agents, and LLM-based multi-agent systems with autonomous AI agents.

\begin{figure}[h!]
     \centering
     \includegraphics[width=0.95\linewidth]{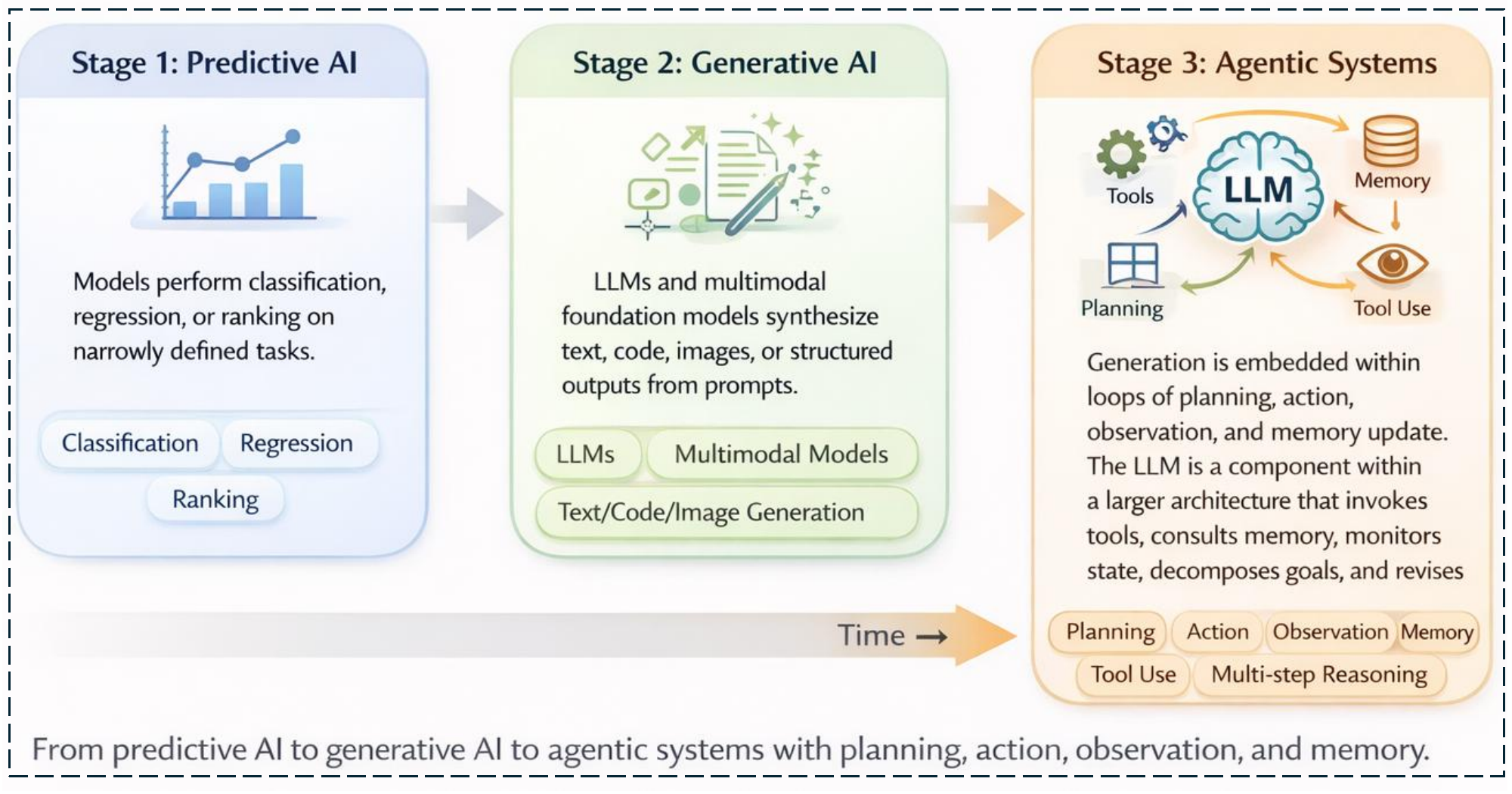}
\caption{A structured evolution of artificial intelligence from predictive models to generative systems and ultimately to agentic architectures integrating planning, action, observation, and memory within autonomous decision-making loops.}
    \label{fig:histry}
\end{figure}

As explained in Figure \ref{fig:histry}, from a historical perspective, the path toward Agentic AI can be understood as a progression through several stages. The first stage emphasized predictive AI, where models performed classification, regression, or ranking on narrowly defined tasks. The second stage introduced generative AI, in which LLMs and multimodal foundation models could synthesize text, code, images, or structured outputs from prompts. The third stage, now rapidly expanding, concerns agentic systems, where generation is embedded within loops of planning, action, observation, and memory update. In this third stage, an LLM is not the entire system, but a component within a larger architecture that may invoke tools, consult memory stores, monitor external state, decompose goals into subtasks, and continuously revise its own strategy. This transition is particularly relevant for scientific, enterprise, robotic, and cyber-physical applications, where decision continuity, contextual awareness, and system-level reliability are indispensable.

Despite substantial progress, the literature remains fragmented. Existing studies frequently examine agent frameworks \cite{garg2025designing, olujimi2025agentic}, autonomous AI assistants \cite{frank2026consumer,kumar2025agentic, luo2025human, bowen2025agentic}, orchestration libraries \cite{piccoli2025large, nygaard2026agentic, chopra2025levels}, local inference engines \cite{park2025survey, gao2026localbench, vake2025hive}, or tool-use paradigms in isolation \cite{li2025review}. Some works focus on LLM reasoning and planning \cite{hao2023reasoning, dou2025plan}, others on memory augmentation \cite{salama2025meminsight, xu2025memory}, retrieval \cite{hong2025enhancing, wang2025leveraging}, or orchestration \cite{luo2025toward, rudra2025composable}, while still others emphasize privacy-preserving local deployment \cite{andreoletti2026privacy, zhang2026privacy}. Yet there is a clear shortage of review-level analyses that jointly study how an inference framework such as \textbf{Ollama} and an agent runtime such as \textbf{OpenClaw} together instantiate a coherent \textbf{Agentic AI Architecture}. This gap is nontrivial. Without a unified architectural perspective, it becomes difficult to answer foundational questions such as: \textbf{What precisely differentiates inference from agency? Why are both layers needed in a practical autonomous system? How should persistent agent runtimes be evaluated when they depend on external model engines? What new safety and governance risks emerge when local inference and autonomous execution are combined? How should future infrastructures evolve toward scalable, trustworthy, and interoperable multi-agent ecosystems?}

The case for studying Ollama is especially strong in the current technological landscape. As concerns about privacy, data governance, regulatory compliance, model cost, and latency continue to grow, local LLM inference has become increasingly attractive. Self-hosted inference engines reduce dependence on external APIs, lower recurring costs in some deployment scenarios, and support offline or edge-based operation \cite{tanikanti2025first, gopee2026self, knoop2026private}. These properties are particularly relevant in domains involving sensitive user data, proprietary enterprise workflows, or constrained connectivity. However, local inference alone does not create an autonomous agent. It provides token generation, not persistent behavior; prediction, not orchestration; language output, not tool-governed action \cite{zhao2025meta, kang2026thunderagent}. Thus, an inference engine such as Ollama should be understood as a necessary but insufficient component of a complete Agentic AI system.

Similarly, the case for studying OpenClaw arises from the growing need for persistent, tool-using, memory-enabled AI runtimes. In contrast to ordinary chat interfaces, OpenClaw-type systems aim to maintain state over time, manage long-term goals, invoke external skills, and operate across multiple channels and contexts \cite{deng2026taming, shan2026don, suwansathit2026systematic}. This aligns closely with the defining properties of autonomous AI agents: persistence, modular actionability, memory continuity, and the ability to reason across multi-step workflows \cite{yu2026agents, alva2026agentic, lee2026toward}. Yet such agent runtimes also introduce new forms of risk. When an LLM is placed inside a persistent execution framework with access to files, APIs, communication tools, or schedulers, the system’s failure modes extend far beyond hallucinated text \cite{cornacchia2025between}. Risks now include memory poisoning, malicious tool invocation, privacy leakage, prompt injection, unsafe automation, and loss of human oversight \cite{he2025emerged, gulyamov2026prompt}. Therefore, OpenClaw is not only architecturally significant but also a compelling case study for examining safety, trust, and governance in full-stack agentic systems.

A full understanding of OpenClaw and Ollama also requires attention to layered architectures. In conventional software engineering, layered design promotes modularity, abstraction, and maintainability by separating system concerns into relatively independent components \cite{thaiya2022software, mari2003impact}. A similar principle applies to Agentic AI. The inference layer is responsible for model execution and token generation \cite{zhang2026toward, brohi2025research}. The runtime layer is responsible for orchestrating agent reasoning loops, memory access, and tool calls \cite{bandi2025rise, alva2026agentic}. The execution layer interfaces with external systems such as browsers, filesystems, APIs, and communication channels \cite{rohitha2025agentic}. The governance layer constrains, monitors, logs, and audits behavior \cite{biswas2026responsible, janakiraman2025governance}. The memory layer preserves continuity across sessions and tasks \cite{sarin2025memoria, kang2025memory}. Distinguishing among these layers is essential for both analysis and design, because the capabilities, bottlenecks, and risks of each layer differ substantially. For example, increasing model size may improve reasoning quality but worsen latency; adding more tools may expand agent utility but enlarge the attack surface; local inference may improve privacy but impose hardware limits. A scientifically rigorous review must therefore consider the entire architecture, rather than conflating all system behavior with the LLM itself.

The importance of this layered view is magnified in the context of LLM-based multi-agent systems. Multi-agent architectures introduce additional complexity, including inter-agent communication, role specialization, distributed memory, coordination protocols, negotiation mechanisms, and conflict resolution strategies \cite{lesser1998reflections}. In such systems, the inference layer may be centralized or distributed, the runtime layer may manage multiple heterogeneous agents, and the governance layer must handle both individual and collective behavior \cite{ahmed2025distributed, alvarez2026explainable}. OpenClaw and Ollama, while often discussed in single-agent deployment contexts, provide conceptual building blocks that can be extended toward broader agent ecosystems. Ollama can support local inference for multiple agents or specialized local models \cite{coelho2025adaptive, liu2025feasibility}, while OpenClaw-like runtimes can coordinate persistent agent roles, memories, and tools \cite{chen2026clawed, weidener2026openclaw, wang2026openclaw}. This creates a fertile research space at the intersection of Agentic AI, multi-agent systems, AI agents, and trustworthy autonomous infrastructures.

Another major motivation for this review arises from the limitations of existing evaluation paradigms. Traditional model-centric benchmarks are insufficient for assessing full-stack agentic systems, as they fail to capture critical properties such as persistence, tool-use reliability, long-horizon planning, recovery behavior, and governance compliance \cite{wang2025ai, yang2025code}. A model that performs strongly on isolated reasoning benchmarks may nonetheless exhibit significant failure modes when deployed within continuous, real-world execution environments \cite{hermawan2025benchmarking, srivastava2025beyondbench, ji2025survey}. Conversely, comparatively smaller or less sophisticated models can demonstrate robust agentic behavior when embedded within architectures that incorporate structured memory, reliable tool abstraction, and well-defined oversight mechanisms \cite{yang2026toward, vishnubhatla2025agentic}. These observations highlight a fundamental shift in the unit of analysis: from isolated model outputs to integrated system-level performance. Consequently, a rigorous investigation of agentic AI must extend beyond model capabilities to encompass architecture, orchestration, safety, governance, and evaluation frameworks in a unified manner.

Against this backdrop, this paper presents a comprehensive and architecturally grounded review of OpenClaw and Ollama within the broader context of Agentic AI. Unlike prior studies that examine inference engines, agent frameworks, or orchestration mechanisms in isolation, this work provides a unified full-stack perspective that explicitly analyzes the interaction between LLM inference systems and persistent agent runtimes. In particular, this paper makes three key contributions. First, it establishes a layered architectural taxonomy of agentic AI systems, distinguishing between inference, runtime, execution, memory, and governance layers. Second, it systematically analyzes Ollama as a local LLM inference layer and OpenClaw as a persistent agent runtime, highlighting their complementary roles in enabling autonomous agent behavior. Third, it integrates a prototype experimental validation of the OpenClaw–Ollama stack, demonstrating that autonomous capabilities such as tool use and persistent memory emerge from system-level integration rather than from model performance alone. Through this combined theoretical and empirical approach, the paper bridges the gap between model-centric AI research and architecture-centric autonomous systems design.

\begin{figure}[h!]
     \centering
     \includegraphics[width=0.75\linewidth]{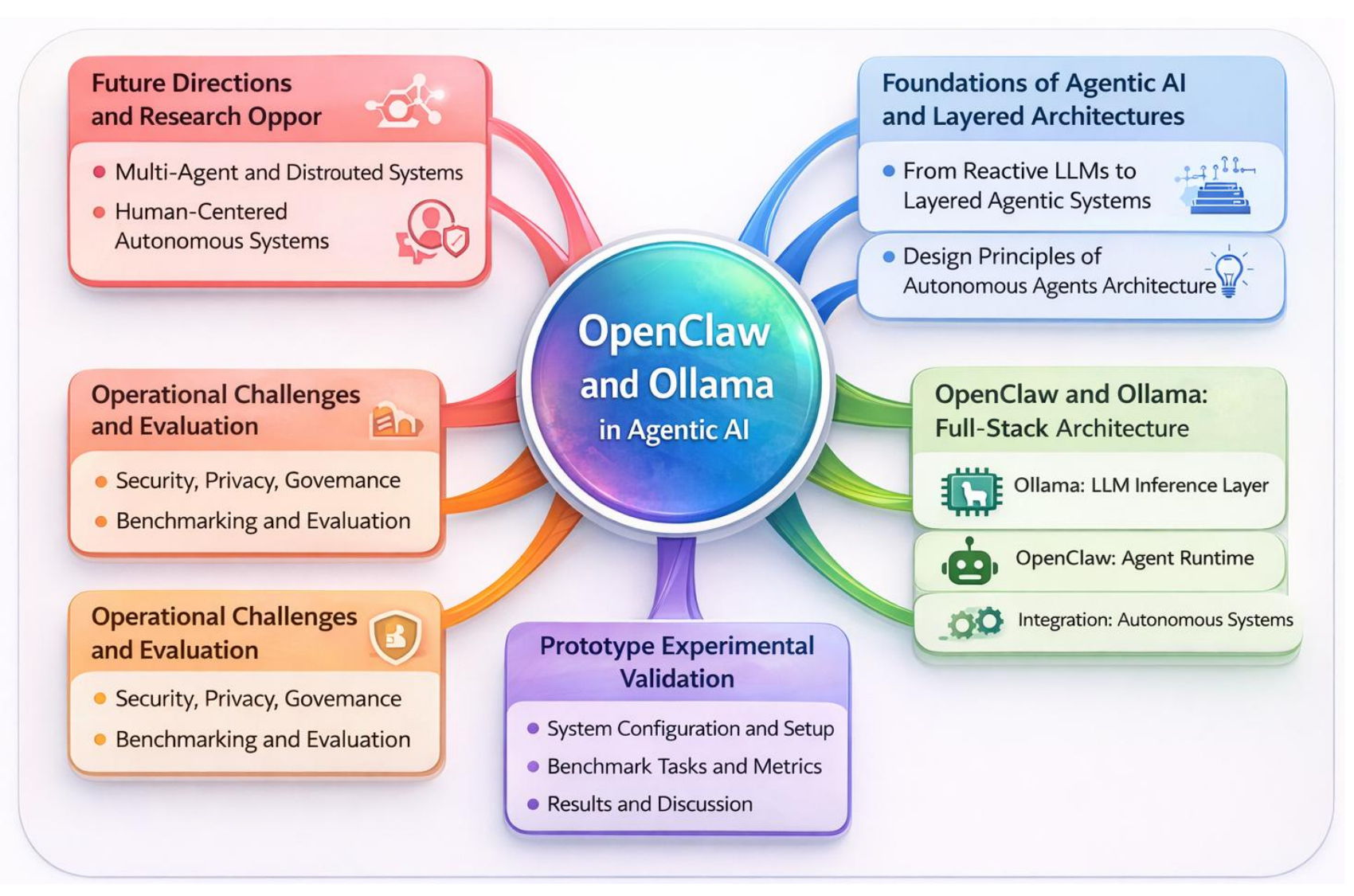}
\caption{Graphical overview of the analytical flow of this review paper. Beginning with the introduction, the study progresses through four interconnected stages: (i) foundational concepts of layered Agentic AI, (ii) architectural integration of Ollama and OpenClaw as a full-stack system, (iii) prototype experimental validation and system-level analysis, and (iv) operational challenges and future directions toward scalable, human-centered, and trustworthy autonomous agent systems.}
    \label{fig:intro_mindmap_openclaw_ollama}
\end{figure}

To provide a clear analytical flow, the review is structured around four major sections, as illustrated in Figure \ref{fig:intro_mindmap_openclaw_ollama}. The first section, \textit{Foundations of Agentic AI and Layered Architectures}, establishes the conceptual basis by tracing the evolution from reactive LLM systems to layered agentic architectures and by defining the core design principles of autonomous agent systems. The second section, \textit{OpenClaw and Ollama: A Full-Stack Agentic AI Architecture}, presents the central technical analysis, examining Ollama as the inference layer, OpenClaw as the runtime layer, and their integration into complete autonomous pipelines. The third section, \textit{Operational Challenges, Safety, and Evaluation}, addresses practical considerations including security, privacy, governance, trust, and benchmarking of persistent agentic systems. The fourth section, \textit{Future Directions and Research Opportunities}, extends the discussion toward scalable multi-agent systems, distributed architectures, and human-centered, responsible AI, identifying critical research gaps and pathways toward trustworthy and interoperable autonomous agent ecosystems.

To make these objectives explicit, this review is guided by the following research questions:

\begin{enumerate}
    \item RQ1: Foundations of Agentic AI and Layered Architectures: How have reactive LLM systems evolved into layered Agentic AI architectures, and what foundational design principles define robust autonomous AI agents and LLM-based multi-agent systems?
   \item  RQ2: OpenClaw and Ollama as a Full-Stack Agentic AI Architecture: How do Ollama and OpenClaw function as complementary inference and runtime layers, and what architectural insights emerge when they are analyzed together as a complete system for Autonomous Agent Systems?
    \item RQ3: Operational Challenges, Safety, and Evaluation: What are the principal security, privacy, governance, reliability, and benchmarking challenges associated with layered Agentic AI Architecture, especially when local inference and persistent autonomous execution are combined?
    \item RQ4: Future Directions and Research Opportunities: What future developments are most critical for advancing scalable, interoperable, human-centered, and trustworthy AI agents, including next-generation LLM-based multi-agent systems with autonomous AI agents?

\end{enumerate}

Figure~\ref{fig:intro_mindmap_openclaw_ollama} synthesizes the conceptual flow of this review by visually integrating the progression from foundational principles of layered Agentic AI to the full-stack architectural analysis of OpenClaw and Ollama, followed by experimental validation, operational challenges, and forward-looking research directions, thereby providing a unified roadmap that bridges theoretical foundations, system-level integration, empirical evaluation, and the future design of scalable and trustworthy autonomous agent systems.

\section{Foundations of Agentic AI and Layered Architectures}

\subsection{From Reactive LLMs to Layered Agentic AI Systems}
\begin{figure}[h!]
     \centering
     \includegraphics[width=0.95\linewidth]{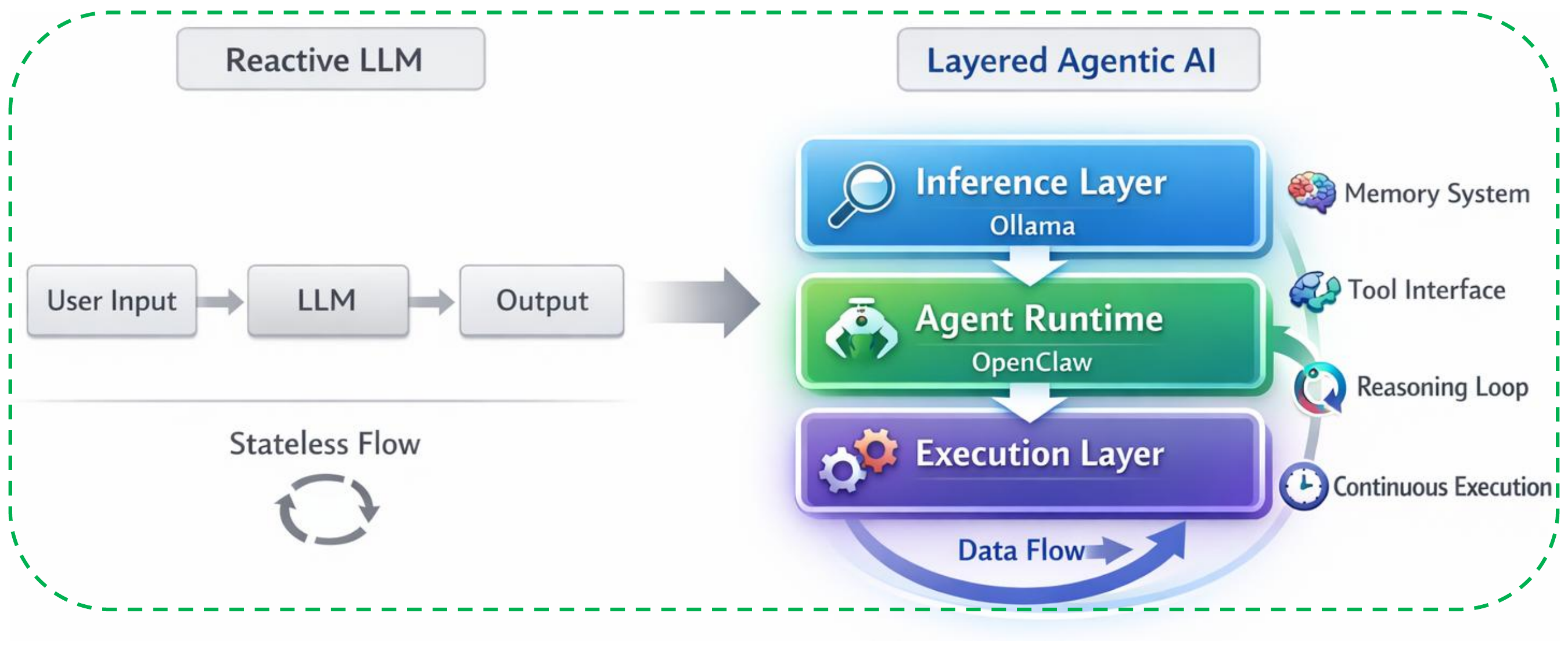}
    \caption{Illustrating the stateless Reactive LLM pipelines with layered Agentic AI systems. The left side shows a Stateless Flow where input is directly mapped to output without memory or action. The right side presents a structured architecture with Inference Layer, Agent Runtime, and Execution Layer, enabling persistent behavior. Key components such as Memory System, Tool Interface, Reasoning Loop, and Continuous Execution support autonomous operation. Data Flow and cyclic feedback illustrate iterative reasoning, highlighting the transition from passive models to fully integrated autonomous agent systems.}
    \label{fig:Foundation1}
\end{figure}

As depicted in Figure \ref{fig:Foundation1}, the evolution of artificial intelligence systems from reactive large LLM interfaces to fully operational Agentic AI systems represents a fundamental paradigm shift in how intelligence is conceptualized, deployed, and evaluated. Traditional LLM applications are inherently stateless and reactive, operating under a prompt-response paradigm in which each interaction is largely independent of previous context unless explicitly provided \cite{chkirbene2024large, chen2025empirical}. While such systems demonstrate strong capabilities in natural language understanding, reasoning, and generation, they lack essential properties required for real-world autonomy, including persistence, long-term memory, goal continuity, and the ability to act upon external environments \cite{gao2025llm, yi2025survey}. Consequently, these systems are limited in handling complex, multi-step tasks that require iterative reasoning, dynamic adaptation, and continuous interaction with tools or environments \cite{li2024survey, hua2025integration}.

In contrast, autonomous agents extend beyond passive response generation by incorporating mechanisms for reasoning, memory, planning, and action within a continuous operational loop \cite{sibai2026path}. These systems are capable of maintaining internal state, tracking objectives over time, and interacting with external resources such as APIs, databases, and physical or virtual tools \cite{kampik2022governance, alva2026agentic}. This transition from reactive models to persistent agents is not merely an incremental improvement but a structural transformation in system design. It reflects a shift from model-centric intelligence toward architecture-centric intelligence, where the behavior of the system emerges from the integration of multiple components rather than from the LLM alone \cite{gupta2026agentic, perwej2026machine}.

A key enabler of this transformation is the introduction of layered AI architecture, in which complex systems are decomposed into distinct but interconnected layers, each responsible for specific functionalities \cite{adabara2025trustworthy}. At a minimum, modern Agentic AI architecture can be conceptualized as consisting of three primary layers as shown in Figure \ref{fig:Foundation1}: (i) the LLM inference layer, responsible for language understanding and generation; (ii) the agent orchestration or runtime layer, responsible for reasoning, planning, and coordinating actions; and (iii) the execution layer, responsible for interacting with external tools, environments, and data sources. This separation of concerns is critical for achieving modularity, scalability, and maintainability in autonomous systems.

Within this layered architectural framework, systems such as Ollama and OpenClaw can be interpreted as representative components operating across distinct strata of the Agentic AI architecture. Specifically, Ollama functions at the LLM inference layer, providing efficient, local, and privacy-preserving model execution capabilities \cite{rodriguez2025privacy, andreoletti2026privacy}. By abstracting model loading, serving, and interaction processes, it enables flexible deployment across local and edge environments \cite{anakabe2026adding}. However, Ollama primarily operates as a stateless inference engine and does not inherently support persistence, memory management, or autonomous decision-making \cite{buyya2026agentic}.

These higher-level capabilities are instead realized at the agent runtime layer, exemplified by OpenClaw, which orchestrates iterative reasoning loops, manages long-term memory, enables tool invocation, and supports continuous task execution \cite{zhu2026tool}. The separation between inference and runtime layers allows independent optimization of system components: inference engines can focus on computational efficiency and model accuracy, while runtime systems are designed to enhance autonomy, planning, and interaction within autonomous agent systems \cite{yang2026openclaw, chen2026inference}.

Several enabling paradigms underpin the functionality of persistent agent systems. One of the most important is the reasoning-action loop \cite{yao2022react, wang2026vla}, often implemented through iterative cycles of thought, action, observation, and memory update \cite{li2026general}. This loop allows agents to decompose complex goals into sub-tasks, evaluate intermediate outcomes, and refine their strategies dynamically \cite{wei2026agentic}. Closely related is the concept of tool calling, where agents invoke external functions or services to extend their capabilities beyond text generation \cite{masterman2024landscape, patil2025berkeley}. This includes web browsing \cite{song2025beyond, pang2025browsemaster}, code execution \cite{guo2024redcode, wang2024executable}, database querying \cite{shorten2025querying, zhang2025agent}, and interaction with domain-specific APIs \cite{elder2026live, ni2025doc2agent}. 

Additionally, memory retrieval mechanisms enable agents to access historical context, past decisions, and accumulated knowledge, thereby overcoming the limitations of fixed context windows in LLMs \cite{zhang2025survey}. Finally, continuous execution mechanisms, such as scheduling or event-driven triggers, allow agents to operate proactively rather than solely in response to user input \cite{iannino2020event, calvaresi2021real}.

Despite their success, stateless LLM systems exhibit several fundamental limitations that motivate the adoption of layered architectures. First, the absence of persistent memory restricts their ability to maintain continuity across interactions, leading to fragmented reasoning and repeated context reconstruction. Second, their reactive nature prevents them from initiating actions or monitoring environments autonomously. Third, their limited integration with external systems constrains their practical utility in real-world workflows. These limitations become particularly pronounced in applications requiring long-horizon planning, multi-step reasoning, or coordination across multiple tools and data sources.

Layered Agentic AI systems address these limitations by decoupling inference from orchestration and execution \cite{kumar2026agentic}. This modular design enables independent scaling and optimization of each layer, facilitating more robust and adaptable systems \cite{bandi2025rise, cao2026auton}. For example, improvements in the LLM inference layer can enhance reasoning quality without altering the runtime logic, while enhancements in the agent runtime can improve planning and execution without requiring retraining of the underlying model \cite{park2026minimizing, li2026large}. Moreover, layered architectures provide a clearer framework for implementing safety mechanisms, monitoring system behavior, and enforcing governance policies across different components \cite{arora2025securing, agarwal2025five}. Table \ref{tab:reactive_vs_agentic} shows the evolution summary from reactive to layered LLM-based agentic AI systems. 

\begin{table*}[ht!]
\centering
\scriptsize
\caption{Evolution from Reactive LLM Systems to Layered Agentic AI Architectures}
\label{tab:reactive_vs_agentic}
\begin{tabular}{p{3cm} p{3.2cm} p{6.2cm}}
\hline
\textbf{System Characteristic} & \textbf{Reactive LLM Systems} & \textbf{Layered Agentic AI Systems} \\
\hline

Interaction Paradigm & Prompt-response, user-driven interaction & Continuous, goal-driven interaction with autonomous execution \\

State Management & Stateless or session-limited context & Persistent memory with long-term context retention \\

Architecture Design & Monolithic LLM-centric pipeline & Layered architecture (Inference, Runtime, Execution) \\

Reasoning Capability & Single-step reasoning per prompt & Multi-step reasoning with iterative refinement loops \\

Execution Ability & No direct action capability & Tool-based action execution in external environments \\

Memory Utilization & Limited to context window & Hierarchical memory (short-term, long-term, episodic) \\

Autonomy Level & Fully reactive & Semi-autonomous to fully autonomous agents \\

Tool Integration & Minimal or absent & Modular tool interfaces with API/function calling \\

Adaptability & Static behavior per prompt & Dynamic planning and adaptation over time \\

System Triggering & User-initiated only & Event-driven and time-triggered execution \\

Scalability & Limited by model and prompt size & Scalable across agents, tools, and distributed systems \\

Deployment Mode & Primarily cloud-based APIs & Local, cloud, or hybrid (e.g., Ollama + OpenClaw) \\

Control Flow & Linear input-output mapping & Cyclic reasoning-action-feedback loops \\

System Intelligence & Model-centric intelligence & Architecture-centric intelligence \\

\hline
\end{tabular}
\end{table*}

\subsection{Design Principles of Autonomous Agents Architecture}
\begin{figure}[h!]
     \centering
     \includegraphics[width=0.75\linewidth]{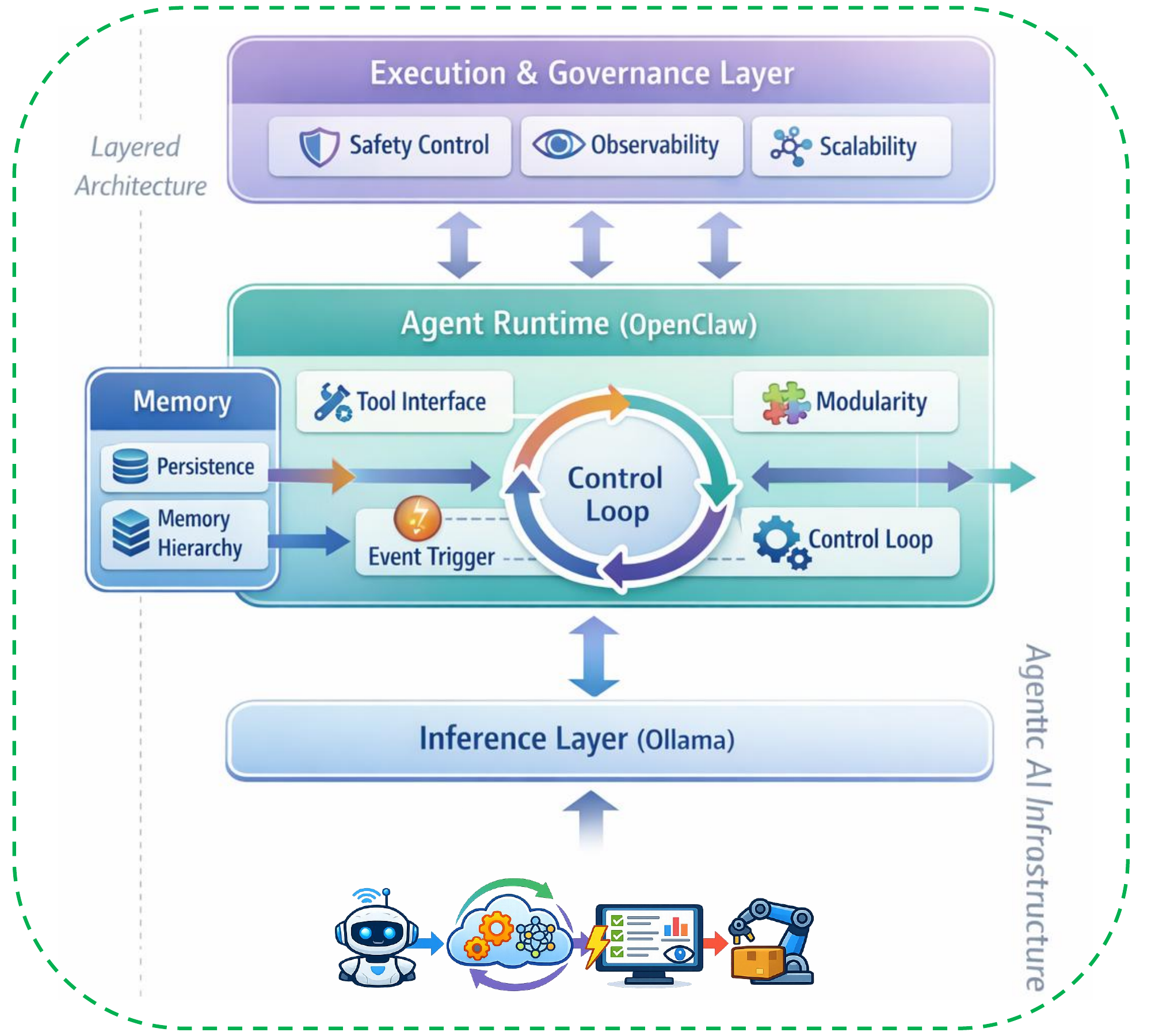}
    \caption{Showing the layered design of autonomous agents architecture, integrating inference (Ollama), runtime (OpenClaw), memory, tools, and governance. Core principles Modularity, Persistence, Memory Hierarchy, Tool Interface, Observability, Safety Control, and Scalability are embedded across layers. Control Loop and Event Trigger mechanisms enable dynamic and continuous execution, while structured Data Flow ensures coordination among components. The diagram emphasizes how layered Agentic AI infrastructure supports reliable, scalable, and secure operation of persistent autonomous agents in real-world environments.}
    \label{fig:Foundation2}
\end{figure}

Robust autonomous agents architecture is fundamentally governed by a set of system-level design principles as shown in Figure \ref{fig:Foundation2},  that extend beyond model capability and into the domain of structured, modular, and controllable AI systems engineering. Unlike monolithic LLM applications, modern Agentic AI systems are inherently composite, integrating multiple interacting subsystems including inference engines, reasoning modules, memory layers, execution pipelines, and governance mechanisms \cite{alva2026agentic, kiasari2026agentic}. The effectiveness of such systems depends critically on how these components are architected, coordinated, and optimized under real-world constraints such as latency, scalability, reliability, and safety \cite{jaggavarapu2025evolution, pati2025agentic}.

One of the most essential principles is modularity, which enables the decomposition of agent systems into independent but interoperable components \cite{rosic2025semantic}. Modular design allows different layers such as the LLM inference layer, the agent runtime layer, and the execution interface to evolve independently without tightly coupling system behavior to a single implementation \cite{xu2025deploying, yazdanian2025comparative}. For example, separating inference (e.g., Ollama) from orchestration (e.g., OpenClaw) enables flexible deployment strategies, where models can be swapped, upgraded, or distributed without redesigning the entire agent pipeline. This abstraction is particularly important in modular AI systems, where heterogeneous components must interoperate across diverse environments and tasks.

Closely related is the principle of persistence, which distinguishes autonomous agents from transient systems \cite{yazdanian2025comparative, astobiza2025ai}. Persistent agents maintain internal state across interactions, enabling long-term goal tracking, contextual continuity, and adaptive behavior \cite{sun2025sophia, deng2025agentic}. This persistence is supported by a structured memory hierarchy, which may include short-term context buffers, long-term storage, episodic memory, and semantic knowledge representations \cite{hu2025memory}. A well-designed persistent memory architecture ensures that relevant information is retained, retrieved, and updated efficiently, allowing the agent to reason over extended temporal horizons while avoiding redundancy or memory saturation \cite{maragheh2025future}.

Another critical design element is tool abstraction, which enables agents to extend their capabilities beyond text generation \cite{lu2025octotools, galster2026configuring}. Tools are typically encapsulated as callable interfaces that allow the agent to interact with external systems such as APIs, databases, file systems, or computational engines \cite{buyya2026agentic}. Effective agent system design requires a clear separation between decision-making logic and tool execution, ensuring that tools are invoked in a controlled, interpretable, and verifiable manner \cite{radanliev2026transparent, hmimou2026dataset}. This abstraction also facilitates extensibility, as new tools can be integrated without modifying core reasoning logic.

Observability is equally important in autonomous systems, as it provides visibility into internal states, decision processes, and execution outcomes \cite{radanliev2026transparent, fournier2025agentic}. Observability mechanisms include logging, tracing, state inspection, and performance monitoring, all of which are essential for debugging, auditing, and optimizing agent behavior \cite{alsayyad2026agenttrace, moshkovich2025beyond}. In the context of Agentic AI infrastructure, observability also supports transparency and trust, enabling users and developers to understand how decisions are made and how actions are executed over time \cite{raheem2025agentic, bandi2025rise}.

To ensure safe and reliable operation, safety constraints must be embedded at multiple layers of the architecture. These constraints may include access control policies, sandboxed execution environments, validation checks for tool outputs, and guardrails for reasoning processes \cite{sharma2025agentic, alqithami2026autonomous}. In persistent systems, safety is not a static property but a dynamic requirement that must be enforced continuously as the agent interacts with evolving environments \cite{kumar2025regulating, leo2026threat}. This necessitates integration of governance mechanisms within the architecture rather than as external add-ons.

Scalability is another key consideration, particularly as agentic systems evolve toward multi-agent and distributed configurations \cite{kim2026cost, chhikara2025mem0}. Scalability involves both horizontal expansion (e.g., multiple agents or distributed inference nodes) and vertical optimization (e.g., efficient memory usage and low-latency execution) \cite{wang2025cognitive, park2025survey}. Architectural decisions such as centralized versus distributed control, and cloud-based versus local inference, directly impact scalability \cite{dhevi2025effective, hossain2025micro}. For instance, centralized architectures simplify coordination but may introduce bottlenecks, whereas distributed architectures enhance robustness and parallelism but require sophisticated synchronization mechanisms \cite{dai2025state, zhao2026survey}.

Control flow within autonomous agents is typically governed by control loops, which define iterative cycles of reasoning, action, observation, and state update \cite{mishra2026sok, wei2026agentic}. These loops may operate in conjunction with event-driven execution, where agents respond to external triggers, and time-triggered autonomy, where actions are initiated based on scheduled intervals \cite{lewis2026quantifying, raeispour2026cyber}. Effective coordination of these mechanisms requires state synchronization across components, ensuring consistency between memory, reasoning state, and external interactions. Without proper synchronization, agents may exhibit inconsistent or erroneous behavior due to stale or conflicting information \cite{maharramova2025artificial, alvarez2026explainable}.

The separation between inference and orchestration layers further enhances system robustness by isolating concerns. The LLM inference layer can be optimized for computational efficiency, model accuracy, and hardware utilization, while the agent runtime layer focuses on decision-making logic, memory management, and execution coordination \cite{park2025survey, park2026minimizing}. This separation also enables hybrid deployment strategies, where inference may occur locally (e.g., via Ollama) while orchestration may be distributed across services or devices.

In summary, as explained in Table \ref{tab:design_principles_agentic}, the design of autonomous agents systems requires a holistic architectural perspective that integrates modularity, persistence, memory hierarchy, tool abstraction, observability, safety, and scalability. These principles collectively define the foundation of modern Agentic AI infrastructure, enabling systems that are not only intelligent but also reliable, adaptable, and controllable. This conceptual framework is essential for analyzing system-level trade-offs and guiding the development of next-generation agentic architectures.

\begin{table*}[ht!]
\centering
\scriptsize
\caption{Core Design Principles and System-Level Roles in Autonomous Agents Architecture}
\label{tab:design_principles_agentic}
\begin{tabular}{p{2cm} p{3cm} p{4cm} p{4.5cm}}
\hline
\textbf{Principle} & \textbf{Definition} & \textbf{Architectural Role} & \textbf{Implementation in Agentic AI Systems} \\
\hline

Modularity & Decomposition of systems into independent, interoperable components & Enables separation of inference, runtime, memory, and execution layers for flexibility and maintainability & Layered architecture with Ollama (inference) and OpenClaw (runtime) decoupled; plug-and-play tool modules \\

Persistence & Ability to maintain state across interactions and time horizons & Supports long-term reasoning, goal tracking, and contextual continuity in autonomous agents & External memory stores, session tracking, and task history integration within runtime layer \\

Memory Hierarchy & Structured organization of short-term and long-term memory & Facilitates efficient retrieval, context compression, and knowledge accumulation & Context buffers, vector databases, episodic logs, and semantic memory modules \\

Tool Abstraction & Encapsulation of external functionalities as callable interfaces & Extends agent capability beyond text generation to real-world interaction & API wrappers, function calling, plugin-based tools (e.g., file systems, web APIs) \\

Observability & Monitoring and tracing of internal states and execution flows & Ensures transparency, debugging capability, and system accountability & Logging pipelines, execution traces, telemetry dashboards, and reasoning logs \\

Safety Control & Enforcement of constraints to prevent unsafe or unintended actions & Maintains secure and trustworthy operation in dynamic environments & Sandboxed execution, access control policies, validation layers, and guardrails \\

Scalability & Ability to expand system capacity across tasks, agents, and infrastructure & Supports multi-agent systems, distributed execution, and workload scaling & Distributed runtimes, parallel agent execution, and scalable inference backends \\

Control Loop & Iterative reasoning-action-feedback execution cycle & Enables adaptive decision-making and continuous refinement of agent behavior & Thought-action-observation loops, feedback integration, and dynamic replanning \\

Event Trigger & External or internal signals initiating agent execution & Allows reactive and proactive system behavior based on environment or schedule & Event-driven pipelines, API triggers, and scheduled task execution (cron/heartbeat) \\

State Synchronization & Consistency of data across memory, reasoning, and execution modules & Prevents conflicts, stale context, and inconsistent agent behavior & Shared memory states, synchronization protocols, and consistency validation mechanisms \\

\hline
\end{tabular}
\end{table*}

\section{OpenClaw and Ollama: A Full-Stack Agentic AI Architecture}
https://github.com/Applied-AI-Research-Lab/OpenClaw-and-Ollama-in-Agentic-AI

\subsection{Ollama as the LLM Inference Layer}

Ollama represents a critical advancement in LLM runtime systems, enabling efficient, local, and privacy-preserving execution of large language models \cite{rodriguez2025privacy}. As a local AI inference engine, Ollama abstracts the complexities of model loading, optimization, and serving into a streamlined interface that allows developers and researchers to deploy on-device AI models with minimal overhead \cite{shahout2026orla, wu2025development}. Unlike cloud-based inference APIs, Ollama operates directly on local hardware, providing deterministic control over model execution, reduced latency, and enhanced data privacy \cite{huang2025llms, rajesh2025production}. This capability is particularly important in applications where sensitive data cannot be transmitted to external servers, such as enterprise workflows, healthcare systems, or edge-based robotic platforms.

\begin{figure}[h!]
     \centering
     \includegraphics[width=0.85\linewidth]{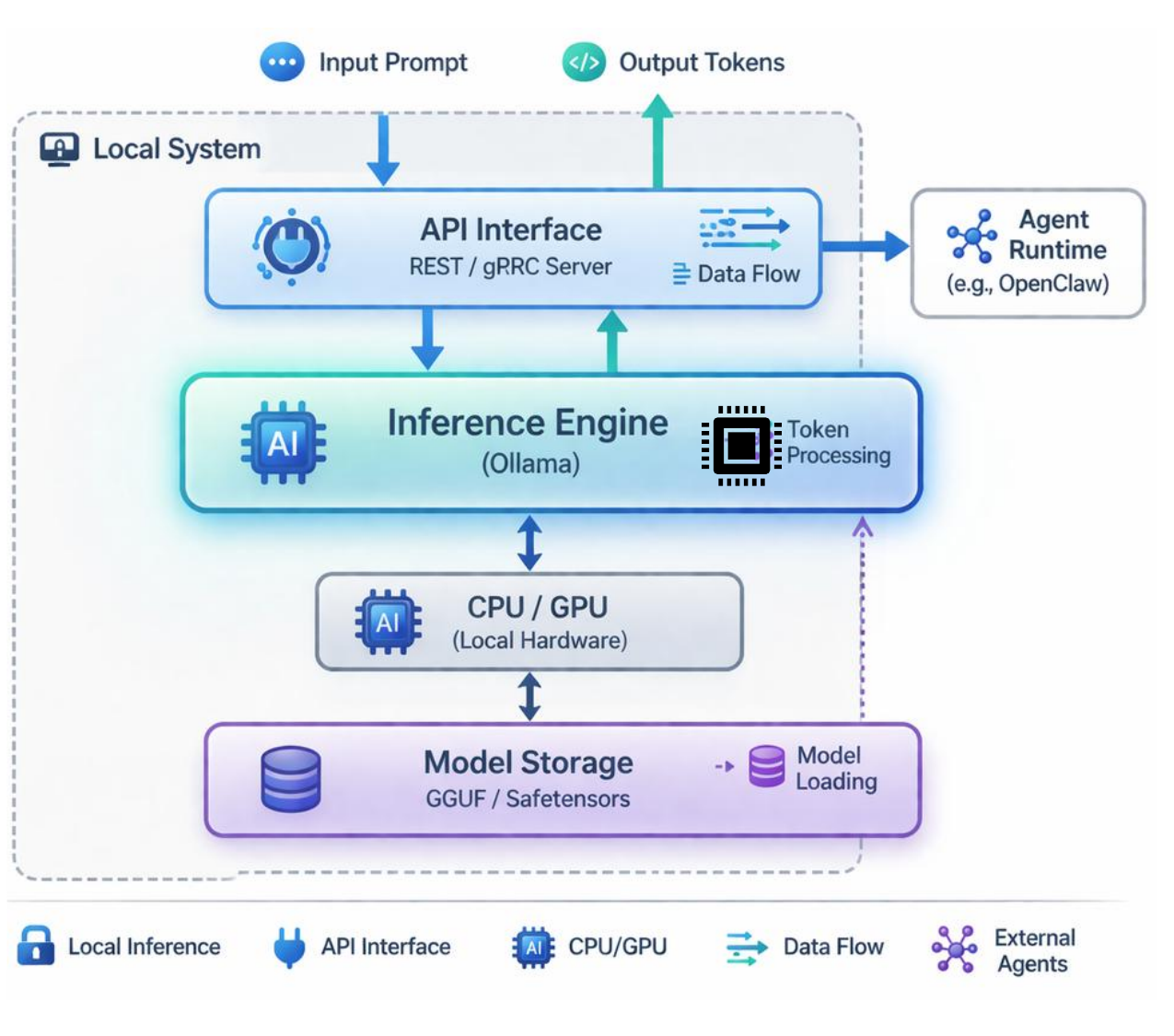}
    \caption{Illustrating a local inference architecture where Model Storage hosts on-device models accessed by the Inference Engine (Ollama) through efficient Token Processing. The API Interface enables communication with external systems and External Agents, while Data Flow arrows depict prompt-to-output transformation. Hardware support via CPU/GPU ensures optimized execution. The Local Inference setup enhances privacy and latency control, positioning Ollama as a foundational component in scalable Agentic AI systems.}
    \label{fig:ollama}
\end{figure}

At its core, as depicted in Figure  \ref{fig:ollama}, the Ollama LLM framework provides a unified interface for managing multiple models, including open-source architectures such as LLaMA, Mistral, and other transformer-based systems \cite{coelho2025adaptive}. It supports model versioning, efficient loading mechanisms, and optimized inference pipelines tailored to available hardware resources, including CPUs and GPUs \cite{marcondes2025using}. Through its lightweight API interface, Ollama allows seamless integration with higher-level systems, enabling agent runtimes such as OpenClaw to invoke LLM capabilities as needed \cite{zhang2026agentic, li2026openclaw}. This API-driven design is essential for modular AI inference architecture, where inference engines operate as independent services within a larger system.

From a systems perspective, Ollama functions as the foundational LLM inference layer within the broader Agentic AI architecture \cite{ke2025survey}. It is responsible for generating context-aware outputs based on input prompts, leveraging pre-trained model weights and token-level processing \cite{menezes2026ai, rosas2025context}. However, it does not inherently maintain state, perform long-term planning, or execute actions \cite{vega2026context, naranjo2026context}. Instead, it operates as a stateless computational engine, producing outputs that must be interpreted and orchestrated by higher-level components \cite{strehlow2026sage, shahout2026orla}. This distinction is crucial: while Ollama provides the cognitive capability of language understanding and reasoning, it does not constitute an autonomous system on its own.

The advantages of local inference through Ollama are explained in Figure \ref{fig:ollama}. First, it eliminates reliance on external cloud services, reducing operational costs and mitigating risks associated with network latency and service availability. Second, it enhances \textbf{data security} by ensuring that sensitive inputs remain within the local environment. Third, it provides \textbf{deterministic performance control}, allowing developers to optimize resource allocation and execution parameters based on specific hardware configurations. These properties make Ollama particularly suitable for edge computing scenarios and environments requiring strict compliance with data governance regulations.

Despite these advantages, Ollama also presents several limitations. The most significant constraint is hardware dependency, as the performance and scalability of local inference are bounded by available computational resources. Large models may require substantial memory and processing power, limiting their applicability on resource-constrained devices \cite{liu2024lightweight, ngo2025edge, zheng2025review}. Additionally, Ollama does not include built-in mechanisms for memory management, task planning, or tool orchestration, which are essential for autonomous behavior \cite{choi2025domain, gong2026structured}. As a result, it must be integrated with higher-level systems to achieve full Agentic AI functionality.

In summary, as illustrated in Figure \ref{fig:ollama}, Ollama serves as a foundational component of AI inference architecture, providing efficient and flexible on-device AI models that enable scalable and privacy-preserving deployments. Its role within the Agentic AI systems stack is to supply the reasoning and language generation capabilities upon which autonomous behavior can be constructed \cite{muraru2025study, bandara2025standardization, bandara2026standardization}. When combined with agent runtimes such as OpenClaw, Ollama becomes a critical enabler of full-stack autonomous agents, bridging the gap between raw model inference and structured, persistent intelligence \cite{weidener2026openclaw}.

\subsection{OpenClaw as the Agent Runtime Layer}

OpenClaw represents a higher-level Agentic AI runtime that transforms stateless LLM outputs into persistent, goal-oriented, and action-capable autonomous agents \cite{chen2026openclaw, chen2026clawed}. Unlike inference engines, OpenClaw is designed as a continuous execution framework that integrates reasoning, memory, tool usage, and scheduling into a unified system \cite{chen2026openclaw}. Its architecture is centered around the concept of persistent autonomy, where agents operate over extended time horizons, maintain internal state, and interact dynamically with external environments \cite{venerito2026agentic, manik2026openclaw}.

\begin{figure}[h!]
     \centering
     \includegraphics[width=0.75\linewidth]{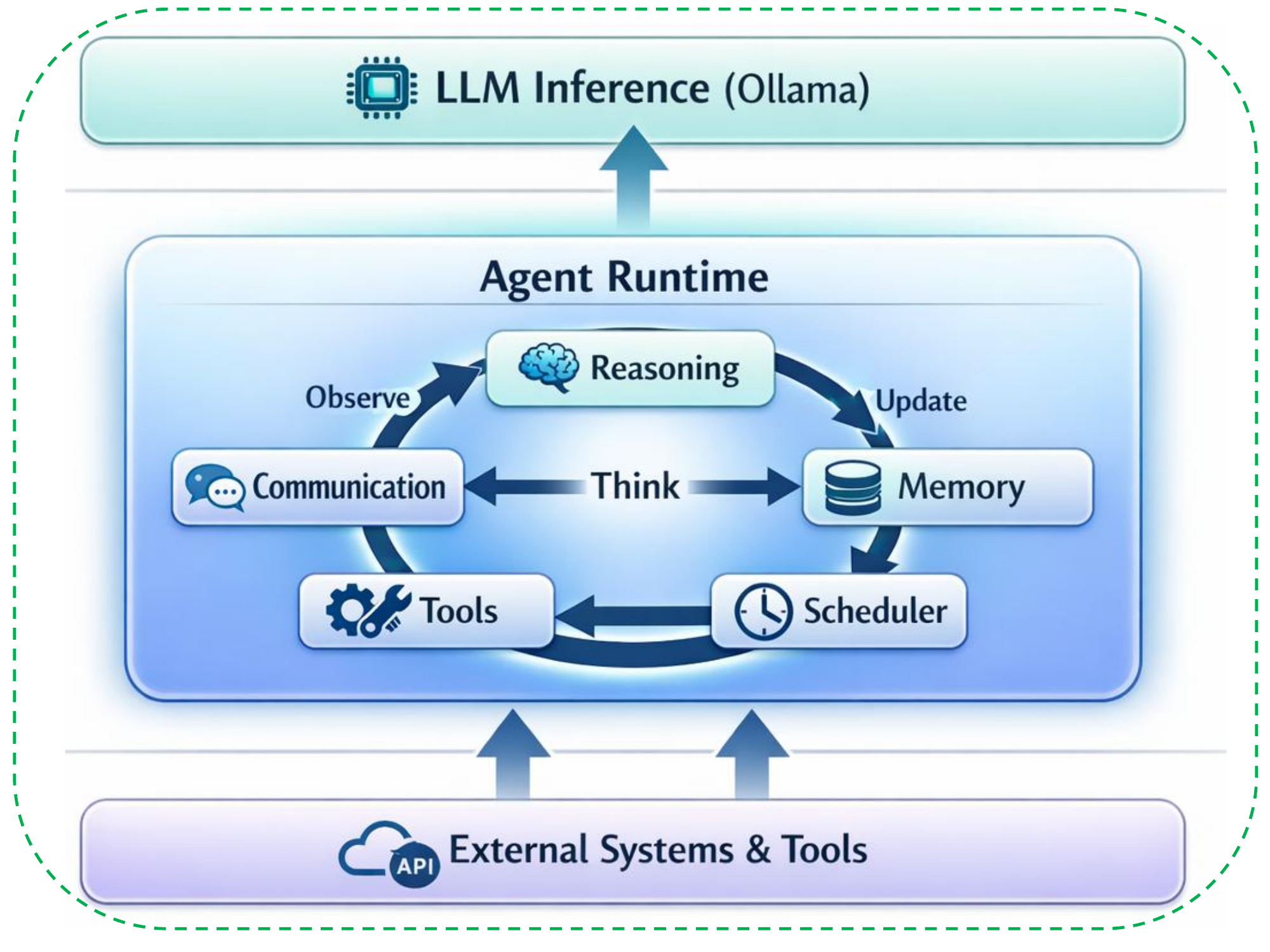}
    \caption{Illustration of AI Agent Runtime coordinating Communication Gateway, Reasoning Module, Memory System, Tool Interface, and Scheduler for persistent autonomous operation. The Reasoning Loop enables iterative observe–think–act cycles, while Continuous Execution supports proactive behavior over time. Integration with LLM Inference provides cognitive outputs, and connections to External Systems enable real-world actions. Structured data flow and feedback loops ensure synchronization between memory, reasoning, and execution within scalable Agentic AI systems.}
    \label{fig:openclaw}
\end{figure}

The OpenClaw architecture consists of several interconnected components as explained in Figure \ref{fig:openclaw}. The communication gateway serves as the interface between the agent and external inputs, including user queries, API calls, or environmental signals \cite{huang2025agentic, buyya2026agentic}. The reasoning module, often powered by an underlying LLM (e.g., via Ollama), interprets inputs and generates plans or actions \cite{marcondes2025using}. The memory system provides both short-term and long-term storage, enabling the agent to retain context, learn from past interactions, and maintain continuity across sessions. The tool execution layer allows the agent to interact with external systems, such as file systems, web services, or domain-specific tools. Finally, the scheduling mechanism ensures continuous operation through event-driven and time-triggered processes.

As shown in Figure \ref{fig:openclaw}, a defining feature of OpenClaw is its use of iterative reasoning loops, in which the agent cycles through stages of observation, reasoning, action, and memory update. This loop enables adaptive behavior, allowing the agent to refine its strategies based on feedback from previous actions \cite{xia2026metaclaw, chen2026clawed}. Unlike reactive systems, which respond only to explicit inputs, OpenClaw supports proactive execution, where agents can initiate actions based on internal goals or scheduled tasks \cite{yang2026openclaw, nie2026synergy}. This capability is essential for applications requiring continuous monitoring, automated workflows, or long-term planning.

The integration of memory is another critical aspect of persistent autonomous agents \cite{kelley2018persistent, liang2025ai}. OpenClaw employs structured memory systems that can store and retrieve information across multiple time scales \cite{chen2026streamingclaw, zhang2026mind}. This includes immediate context for ongoing tasks, as well as long-term knowledge that informs future decisions. By externalizing memory from the LLM, OpenClaw overcomes the limitations of fixed context windows and enables scalable, stateful operation.

Tool abstraction further enhances the capabilities of OpenClaw by enabling interaction with external resources \cite{li2026defensible, ding2026automating, suwansathit2026systematic}. Tools are encapsulated as modular components that can be invoked by the agent during execution \cite{roman2026orchestral, xu2026evolution}. This modularity allows for extensibility and customization, enabling the system to adapt to diverse application domains \cite{dakov2025event, alva2026agentic}. Combined with its scheduling and orchestration mechanisms, OpenClaw forms a comprehensive AI agent orchestration platform capable of coordinating complex workflows \cite{roman2026orchestral}.

Importantly, OpenClaw operates above the LLM inference layer, relying on systems such as Ollama to provide language understanding and reasoning capabilities. This separation of concerns allows OpenClaw to focus on higher-level functions such as planning, memory management, and execution coordination. It also enables flexibility in deployment, as different inference engines can be integrated without altering the core runtime architecture.

However, the increased autonomy and complexity of OpenClaw introduce new challenges. Persistent systems must address issues related to security, reliability, and governance, particularly when interacting with external tools and environments. Ensuring safe and controlled operation requires robust mechanisms for monitoring, validation, and error handling. Additionally, the integration of multiple components increases system complexity, necessitating careful design and optimization.

\subsection{Integration: From Inference to Autonomous Agent Systems}

\begin{figure}[ht!]
     \centering
     \includegraphics[width=0.95\linewidth]{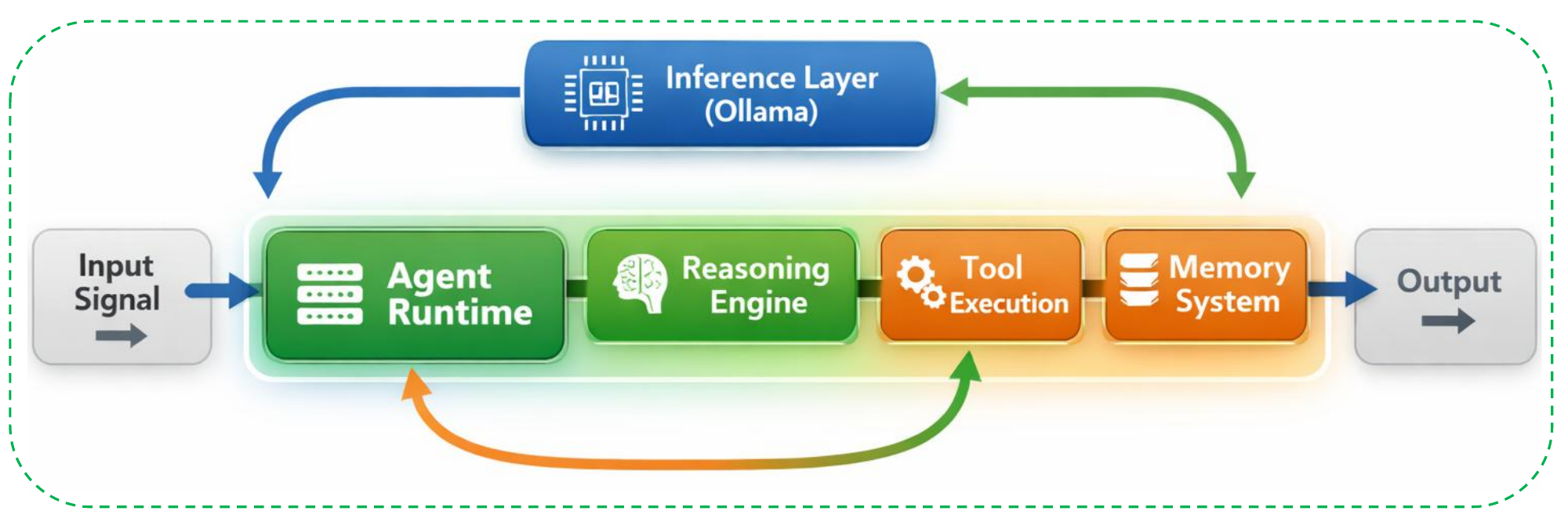}
    \caption{OpenClaw–Ollama Integration: From Inference to Autonomous Agent Systems.
The diagram illustrates the end-to-end Agentic AI stack, where an \textbf{Input Signal} is processed by the Agent Runtime (OpenClaw), which manages task context and orchestration. The request is forwarded to the Inference Layer (Ollama) for language reasoning, producing outputs consumed by the Reasoning Engine. Based on this, the system performs Tool Execution in external environments. Results are stored in the Memory System and fed back through Feedback Loops, ensuring continuous adaptation and structured Data Flow across the autonomous agent pipeline.}
    \label{fig:integration}
\end{figure}

The integration of Ollama and OpenClaw represents a canonical realization of a layered Agentic AI stack as illustrated in Figure \ref{fig:integration}, where LLM + agent runtime integration enables the transformation of stateless model inference into persistent, goal-driven autonomous agent systems architecture. In this unified framework, Ollama operates as the LLM inference layer, providing token-level reasoning and language generation, while OpenClaw functions as the agent runtime layer, orchestrating planning, memory management, tool invocation, and continuous execution. The synergy between these layers establishes a full-stack system capable of perception, reasoning, action, and adaptation over extended time horizons.

At a system level, the integrated pipeline follows a structured data and control flow. The process begins with an input signal, which may originate from a user, an external API, or an event-driven trigger. This input is first routed through the OpenClaw communication gateway, where it is contextualized using available memory and task history. The processed input is then forwarded to the Ollama inference engine, which generates a response based on the current prompt and context. This response is not treated as a final output but as an intermediate reasoning artifact, which is interpreted by OpenClaw’s reasoning module.

Subsequently, OpenClaw evaluates the LLM output to determine whether it represents a final response, a plan, or an actionable command. If action is required, the runtime layer invokes appropriate tools through its execution pipeline, interacting with external systems such as databases, file systems, or web services. The outcomes of these actions are then fed back into the system, updating the memory store and potentially triggering additional reasoning cycles. This iterative loop comprising inference, reasoning, action, and memory update continues until the task is completed or a termination condition is met. Thus, the integration enables a closed-loop system that supports both reactive and proactive behaviors.

\begin{table}[t]
\centering
\caption{Comparison of Monolithic LLM Systems and Layered OpenClaw–Ollama Agentic AI Stack}
\label{tab:integration_comparison}
\begin{tabular}{p{3cm} p{4cm} p{5cm}}
\hline
\textbf{Aspect} & \textbf{Monolithic LLM Systems} & \textbf{Layered Agentic AI Stack (Ollama + OpenClaw)} \\
\hline
Architecture & Single unified model pipeline & Modular layered architecture \\
State Management & Stateless or session-limited & Persistent memory across tasks \\
Execution Capability & No direct action execution & Tool-based external interaction \\
Scalability & Limited by model design & Independent scaling of layers \\
Deployment & Cloud-dependent & Local, hybrid, or distributed \\
Latency Control & External API dependent & Locally optimized inference \\
Adaptability & Fixed behavior & Dynamic planning and adaptation \\
Safety Control & Limited guardrails & Multi-layer governance mechanisms \\
\hline
\end{tabular}
\end{table}

As summarized in Table~\ref{tab:integration_comparison}, the transition from monolithic systems to a layered Agentic AI architecture introduces significant improvements in modularity, scalability, and functional capability. In monolithic systems, inference, reasoning, and output generation are tightly coupled within a single model, limiting flexibility and extensibility. In contrast, the OpenClaw Ollama integration decouples these concerns, allowing each layer to be optimized independently. For instance, inference performance can be enhanced through hardware acceleration or model selection within Ollama, while orchestration logic can be refined within OpenClaw without modifying the underlying model.

Despite these advantages, integration also introduces several technical challenges. One of the primary concerns is latency, as the interaction between inference and runtime layers may involve multiple sequential steps, including prompt construction, model invocation, tool execution, and memory updates. Efficient pipeline design and caching strategies are therefore essential to maintain real-time responsiveness. Another challenge is synchronization, particularly in scenarios involving concurrent tasks or multi-agent coordination. Ensuring consistent state across memory, reasoning modules, and external systems requires robust synchronization mechanisms and conflict resolution strategies.

Memory consistency is another critical issue, as the correctness of agent behavior depends on accurate and up-to-date contextual information. Inconsistent or outdated memory can lead to erroneous decisions or redundant actions. This necessitates careful design of memory retrieval, update, and validation processes. Additionally, the tool invocation pipeline must be designed to ensure reliability, security, and interpretability. Each tool call introduces potential points of failure, requiring validation checks, error handling, and fallback mechanisms.

From an architectural perspective, the layered approach offers several benefits. It enables modular upgrades, where improvements in one layer do not necessitate changes in others. It supports flexible deployment, allowing systems to operate locally, in the cloud, or in hybrid configurations. It also facilitates system-level optimization, where performance can be tuned across layers rather than within a single monolithic model. These properties are particularly important for scaling autonomous agent systems to complex, real-world applications.

Compared to monolithic agent systems, the OpenClaw–Ollama stack provides a more transparent and controllable architecture. By exposing intermediate states and interactions, it enhances observability and enables fine-grained control over system behavior. This is essential for debugging, auditing, and ensuring compliance with safety and governance requirements. Furthermore, the separation of inference and orchestration aligns with established principles in distributed systems and software engineering, reinforcing the robustness and extensibility of the architecture.

In conclusion, the integration of Ollama and OpenClaw exemplifies a scalable and modular Agentic AI stack that bridges the gap between model inference and autonomous system behavior. By combining the strengths of local LLM inference with persistent agent orchestration, this architecture enables the development of advanced tool-using AI systems capable of continuous reasoning, adaptive planning, and real-world interaction. This layered approach not only enhances system capability but also provides a clear framework for addressing challenges in performance, reliability, and governance.

\section{Prototype Experimental Validation of the OpenClaw-Ollama Full-Stack Agentic AI Architecture}
\label{sec:validation}

To ground the architectural discussion in reproducible empirical evidence, this section presents a controlled systems-level evaluation of the integrated OpenClaw--Ollama stack. The objective is not to claim state-of-the-art benchmark superiority over existing agent systems, but to demonstrate, under transparent and reproducible conditions, how full-stack Agentic AI behavior differs across three progressively richer system configurations. The central hypothesis driving this evaluation is that the emergent capabilities of an autonomous agent specifically the ability to access structured local data, invoke external tools, and maintain coherent multi-turn memory arise from the coordinated interplay between the LLM inference layer and the OpenClaw agent runtime, rather than from the language model alone.
This hypothesis manifests in a concrete and testable prediction: a monotonically increasing task success rate from the raw inference baseline (C1), through the runtime-enabled but stateless configuration (C2), to the full persistent-memory configuration (C3). If this ordering holds across multiple models and task categories, it provides direct empirical support for the architectural thesis presented throughout this paper.
The experiment is designed to be lightweight, requiring no third-party crowdsourced annotation and no cloud-hosted infrastructure, while remaining sufficiently rigorous to quantify measurable differences between configurations. All benchmarking code, task definitions, evaluation harness, and logging infrastructure are released as an open-source repository to maximize reproducibility and to serve as a reference implementation for the agentic AI research community.

\noindent\textbf{Code and Data Availability:} All experimental code, datasets, configuration files, evaluation scripts, and detailed results are publicly available at \url{https://github.com/Applied-AI-Research-Lab/OpenClaw-and-Ollama-in-Agentic-AI}.

\subsection{Experimental Configuration and System Environment}
\label{sec:setup}

The experimental environment consists of a single-node server running Ubuntu 22.04 with an NVIDIA H100 PCIe 81\,GB GPU. Ollama (the LLM inference server) and OpenClaw (version 2026.4.5, the agent runtime) are both deployed in \textit{host mode}, meaning both processes run natively on the operating system without container isolation. This choice eliminates inter-process networking latency from the latency measurements and reflects a common deployment pattern for enterprise-local agentic systems. The model-serving endpoint is exposed on \texttt{localhost:11434} (Ollama) and the OpenClaw gateway API on \texttt{localhost:18789}. The evaluation harness is implemented in Python~3.12 inside an isolated virtual environment and communicates with Ollama via its native REST API and with OpenClaw via its command-line interface.

Two open-weight language models are evaluated to assess whether the observed configuration-level trends are model-specific or architectural in nature:

\begin{itemize}
  \item \textbf{Qwen3.5:4b} -- a 4-billion-parameter model from the Qwen3.5 series~\citep{qwen3_5_4b_ollama}, representative of compact but capable instruction-tuned models with strong arithmetic reasoning.
  \item \textbf{Gemma4:e4b} -- a 4-billion-parameter model from Google's Gemma 4 series~\citep{gemma4_e4b_ollama}, a recent open-weight model optimized for instruction following and tool-augmented inference.
\end{itemize}

Both models are loaded directly into Ollama at 4-bit precision and share identical inference parameters (temperature 0.2, no top-p sampling) across all three experimental configurations. The use of two structurally different models enables a cross-model comparison that tests whether the capability progression from C1 to C3 is model-agnostic, a key requirement for any architectural claim to be considered generalizable rather than model-specific. Each configuration is evaluated over three independent repetitions per model; all metrics are reported as mean\,$\pm$\,standard deviation across repetitions.

The three experimental configurations are defined as follows:

\begin{itemize}
  \item\textbf{C1 -- Ollama-only baseline.} The model is queried directly through Ollama's native API with no tool availability, no persistent memory, and no session continuity. Local file references in task prompts are passed as-is (the raw relative path string); the evaluation harness applies file-content enrichment only when the OpenClaw orchestration layer is active (C2 and C3). Under C1, the model therefore receives an unresolvable path and cannot access document content. This configuration isolates the contribution of pure LLM inference.
  \item\textbf{C2 -- OpenClaw stateless.} The task is routed through OpenClaw's agent runtime, making the full toolset available (file read/write, arithmetic, external API queries, cron scheduling). Critically, between each task, the evaluation harness performs a full stateless reset: session files, the SQLite memory store, and all workspace markdown memory files are wiped, ensuring that no information can persist across task boundaries. This configuration isolates the contribution of tool-calling capability without persistent memory.
  \item\textbf{C3 -- OpenClaw persistent (full stack).} Identical to C2 in terms of the available toolset and harness file-injection. However, persistent memory is fully enabled: each task is issued a per-task session identifier (ensuring a fresh chat context per task), but the on-disk memory layer including the workspace markdown journal (\texttt{MEMORY.md}), the SQLite episodic store, and goal-tracking files accumulates unmodified throughout the entire evaluation run. Tasks can therefore retrieve context committed to the persistent store by earlier tasks, enabling genuine long-horizon behaviour.
\end{itemize}

\subsection{Benchmark Task Design}
\label{sec:tasks}

The benchmark consists of 15 tasks partitioned into three functional categories, each targeting a distinct capability layer of the Agentic AI stack. Task definitions, scoring rubrics, and supporting data files are all committed to the public repository.

\textbf{Reactive tasks (3 tasks)} evaluate whether the system can extract specific factual information from a structured local text document (\texttt{reactive\_notes.md}), which contains a short project brief comprising a codename, a deadline, an owner, a priority level, and a blocker. The three tasks ask for the project codename, the project deadline, and the project owner respectively. Under C1, the model receives only the raw file path as a string; the evaluation harness applies file-content enrichment only when the OpenClaw orchestration layer is active (C2 and C3), causing C1 queries to receive an unresolvable relative path rather than the document text. Under C2 and C3, the harness injects the full document text into the prompt prior to submission. Correctness requires an exact or regex-matched string extraction.

\textbf{Tool-use tasks (6 tasks)} evaluate higher-order operational capabilities: (T4) compute the total order cost from a three-row CSV (\texttt{tool\_data.csv}) by multiplying each item's quantity by its unit price and summing; (T5) identify the item with the highest quantity in the same CSV and return its name exclusively; (T6) read the reactive notes document, generate a three-bullet summary, and write it to a new file (\texttt{generated\_summary.md}), with success verified both by content and by file existence; (T11) call the local Ollama REST API at \texttt{/api/tags} and return a fixed confirmation string if the response contains a top-level \texttt{models} key; (T12) evaluate a compound arithmetic expression $((12.5 \times 4) + (3.2 \times 5))$ with an exact numeric answer; and (T13) generate a valid cron expression for a task that runs at 09:00 daily.

\textbf{Persistent tasks (6 tasks)} evaluate the agent's ability to store and retrieve information across task boundaries. The six tasks form three store--recall pairs. T7 instructs the agent to remember a formatting preference (future summaries must use bullet points with no introductory text); T8 asks the agent to use that preference to summarise the reactive notes document in exactly two bullet points. T9 instructs the agent to remember an alias mapping (project alias is ``LR''); T10 asks the agent to retrieve and return only the stored alias. T14 is a \textit{memory-update probe}: issued after T9 has stored ``LR'', it informs the agent that the alias has been updated to ``PR'' and asks it to return the \textit{current} alias testing whether explicit in-prompt updates correctly override stale memory. T15 is a \textit{style-recall probe}: it asks the agent to describe, without any document reference, the formatting preference it was asked to remember, testing unprompted retrieval of opaque stored preferences. Under C2, the full stateless reset between tasks means no recall task can retrieve information from its paired store task, providing a controlled memory-isolation negative test. Under C3, the accumulating on-disk memory enables all six tasks.

Each task is scored using one or more complementary criteria: \texttt{expected\_contains} (all listed tokens must appear in the response), \texttt{expected\_exact} (exact string match after normalization), \texttt{expected\_regex} (a case-insensitive regular expression match allowing for paraphrastic variation in acknowledgment phrasing), \texttt{expected\_not\_contains} (rejection of known failure phrases such as \textit{``unable to''} or \textit{``hard constraint''}), and \texttt{verify\_file\_exists} (physical verification that a specified output file was created on disk). This multi-criterion scoring approach reduces false negatives arising from superficially different but semantically equivalent model responses.

\begin{tcolorbox}[
colback=green!6,
colframe=blue!60!black,
title={Algorithm 1: Prototype Experimental Validation of the OpenClaw--Ollama Full-Stack Agentic AI Architecture},
fonttitle=\bfseries,
sharp corners,
boxrule=0.8pt,
left=2mm,right=2mm,top=1mm,bottom=1mm
]
\footnotesize
\textbf{Input:} Task set $\mathcal{T}=\{t_1,t_2,\dots,t_n\}$, local LLM hosted in Ollama, OpenClaw runtime, toolset $\mathcal{U}$, memory store $\mathcal{M}$, experimental configurations $\mathcal{C}=\{c_1,c_2,c_3\}$\\
\textbf{Output:} System-level evaluation results across configurations

\begin{enumerate}
    \item Initialize three configurations:
    \begin{enumerate}
        \item $c_1$: Ollama-only baseline (no tools, no memory, stateless)
        \item $c_2$: OpenClaw + Ollama with tools; full stateless reset per task (session, SQLite, workspace memory files)
        \item $c_3$: Full OpenClaw + Ollama; per-task session identifier (fresh chat context per task) with accumulating on-disk memory (\texttt{MEMORY.md}, SQLite, goal files)
    \end{enumerate}

    \item Partition task set $\mathcal{T}$ into:
    \begin{enumerate}
        \item Reactive tasks: local document lookup
        \item Tool-use tasks: file I/O, computation, API calls, scheduling
        \item Persistent multi-turn tasks: memory store and recall
    \end{enumerate}

    \item For each configuration $c \in \mathcal{C}$:
    \begin{enumerate}
        \item Deploy the corresponding inference-runtime pipeline
        \item Reset logs, counters, and memory state as required by $c$
    \end{enumerate}

    \item For each task $t_i \in \mathcal{T}$:
    \begin{enumerate}
        \item Receive task input through the communication interface
        \item If $c = c_2$: perform full stateless reset (session files, SQLite rows, workspace markdown memory)
        \item If $c \in \{c_2, c_3\}$: route prompt through OpenClaw orchestration interface
        \item Else if $c = c_1$: forward prompt directly to Ollama REST API
        \item Forward contextualized prompt to Ollama for local inference
        \item Receive model output $o_i$; extract structured answer field
        \item If tool use is indicated, invoke tool $u \in \mathcal{U}$ through OpenClaw
        \item Observe execution result; update memory store $\mathcal{M}$ if $c = c_3$
        \item Re-enter reasoning loop until termination condition is satisfied
        \item Log latency, tool calls, memory events, and final outcome
    \end{enumerate}

    \item Compute evaluation metrics for each configuration:
    \begin{enumerate}
        \item Task success rate, tool-call accuracy, memory recall accuracy
        \item End-to-end latency, reasoning iteration count
    \end{enumerate}

    \item Compare $c_1$, $c_2$, $c_3$ to quantify the contribution of each architectural layer

    \item Analyze trace logs to identify failure modes and architectural bottlenecks
\end{enumerate}
\end{tcolorbox}

\subsection{Evaluation Metrics}
\label{sec:metrics}

Six system-level metrics are computed for each configuration and model. \textit{Task success rate} is the primary metric, defined as the fraction of tasks whose output satisfies all applicable scoring criteria. \textit{Tool-call accuracy} is the task success rate restricted to the six tool-use tasks, providing a focused measure of the agent's tool-using capability. \textit{Memory recall accuracy} is defined analogously for the six persistent tasks. \textit{Average latency} measures wall-clock end-to-end time per task in milliseconds, from prompt submission to structured answer extraction. \textit{Average reasoning steps} counts the number of inference--orchestration iterations per task, serving as a proxy for the depth of the reasoning process engaged by the runtime. Each run generates a structured JSON trace recording the full prompt, model response, tool calls, memory events, scoring breakdown, and timing; these traces constitute the primary audit artefact and are archived alongside the evaluation results.

\subsection{Results}
\label{sec:results}

\begin{figure}[t]
  \centering
  \includegraphics[width=0.72\linewidth]{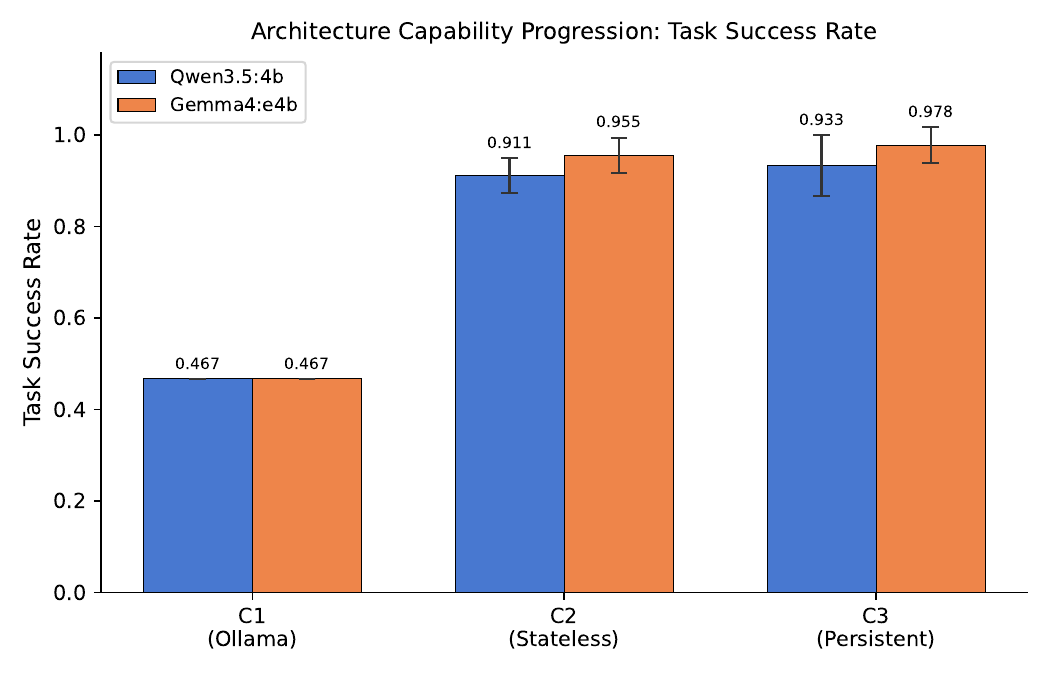}
  \caption{Architecture capability progression: task success rate for Qwen3.5:4b and Gemma4:e4b across configurations C1 (Ollama baseline), C2 (OpenClaw stateless), and C3 (OpenClaw persistent). Both models exhibit a strictly monotonically increasing progression ($\text{C1} < \text{C2} < \text{C3}$), confirming that each architectural layer contributes measurably and independently to overall system performance.}
  \label{fig:progression}
\end{figure}

\begin{figure}[t]
  \centering
  \includegraphics[width=0.72\linewidth]{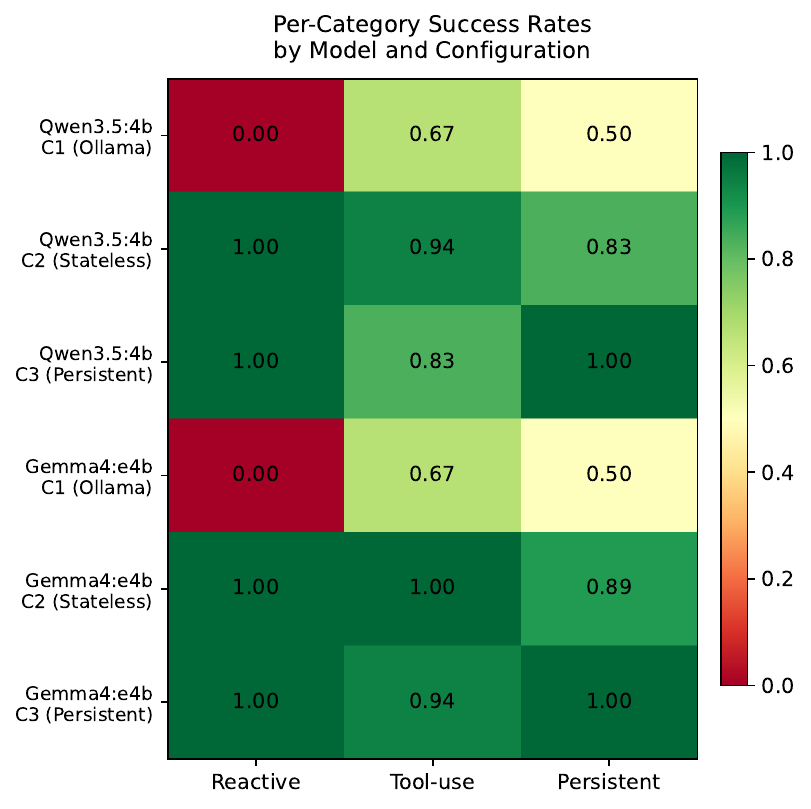}
  \caption{Per-category success rate heatmap ($n=3$ reps, mean) across all model--configuration pairs. Rows represent model--configuration combinations; columns represent task categories (Reactive, Tool-use, Persistent). Colour encodes success rate from red (0.00) through yellow to green (1.00). The gradient reveals that reactive capability is fully absent under C1 and fully present under C2 and C3, while persistent scores reach 1.00 exclusively under C3.}
  \label{fig:heatmap}
\end{figure}

\begin{figure}[t]
  \centering
  \includegraphics[width=0.92\linewidth]{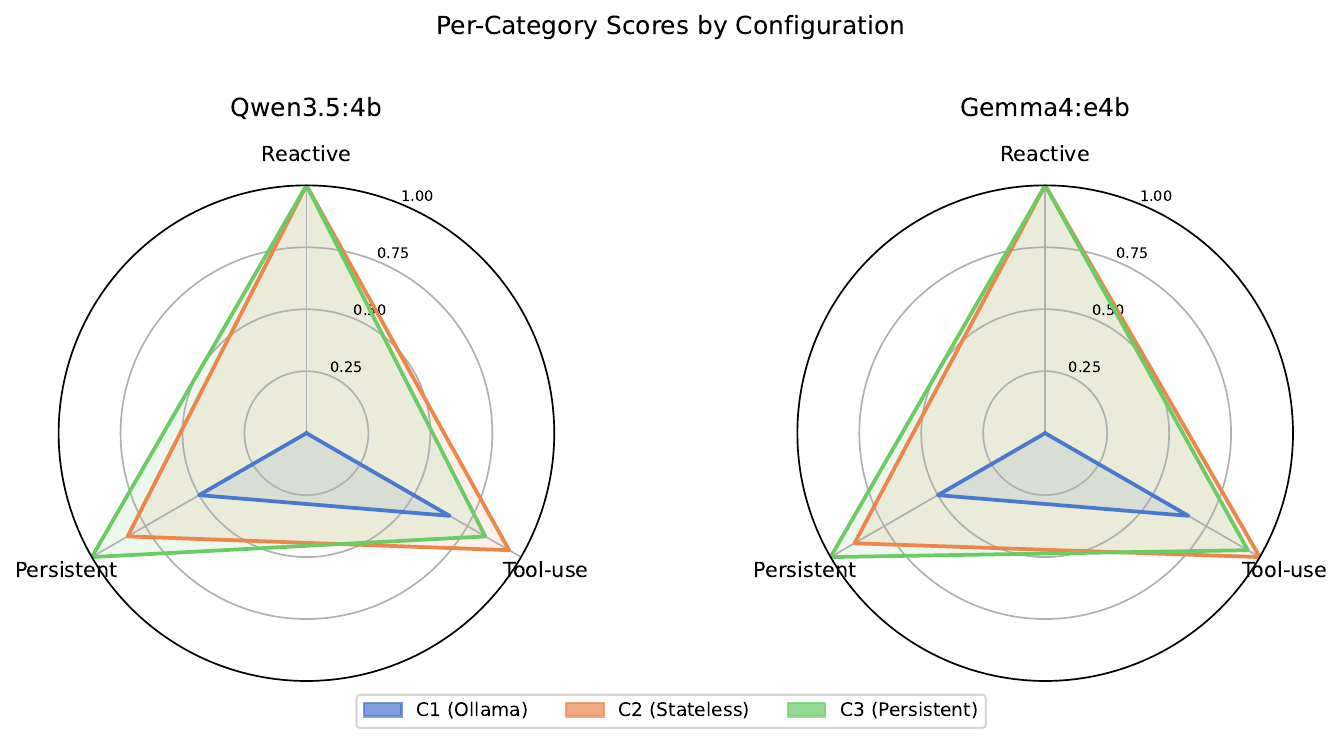}
  \caption{Side-by-side radar charts of per-category scores for Qwen3.5:4b (left) and Gemma4:e4b (right). Each chart overlays C1 (blue), C2 (orange), and C3 (green) across the three task categories. The Gemma4:e4b C2 polygon reaches full coverage on the tool-use vertex (1.000), whereas the Qwen3.5:4b C3 polygon shows a visible dip on tool-use (0.833) due to T11 routing failures the only systematic inter-model difference across all configurations.}
  \label{fig:radar}
\end{figure}

Table~\ref{tab:main-results} reports the primary quantitative results for both models across all three configurations, aggregated over three independent repetitions ($n=3$) per model. Figure~\ref{fig:progression} visualises the task success rate progression from C1 through C3; Figure~\ref{fig:heatmap} provides a per-category success rate heatmap for all model--configuration pairs; and Figure~\ref{fig:radar} shows the same data as side-by-side radar charts, one per model, enabling a direct cross-model comparison of category-level capability profiles.

\begin{table}[t]
\centering
\caption{Prototype validation results ($n=3$ reps, mean\,$\pm$\,std, 15 tasks). C1 is the raw Ollama baseline; C2 adds OpenClaw tool-calling with per-task stateless memory reset; C3 enables full persistent on-disk memory and tools. Bold values indicate the best result per column for each model.}
\label{tab:main-results}
\begin{tabular}{llrrrrr}
\toprule
Model & Config & Success & Tool Acc & Mem Acc & Latency\,(ms) & Pass/Total \\
\midrule
\multirow{3}{*}{Qwen3.5:4b}
 & C1 (Ollama only)         & 0.467{\tiny$\pm$0.000} & 0.667{\tiny$\pm$0.000} & 0.500{\tiny$\pm$0.000} & \textbf{967{\tiny$\pm$91}}    & 7.0/15 \\
 & C2 (OpenClaw, stateless) & 0.911{\tiny$\pm$0.038} & \textbf{0.944{\tiny$\pm$0.096}} & 0.833{\tiny$\pm$0.000} & 11{,}969{\tiny$\pm$695}        & 13.7/15 \\
 & C3 (OpenClaw, persistent)& \textbf{0.933{\tiny$\pm$0.067}} & 0.833{\tiny$\pm$0.167} & \textbf{1.000{\tiny$\pm$0.000}} & 11{,}630{\tiny$\pm$422} & \textbf{14.0/15} \\
\midrule
\multirow{3}{*}{Gemma4:e4b}
 & C1 (Ollama only)         & 0.467{\tiny$\pm$0.000} & 0.667{\tiny$\pm$0.000} & 0.500{\tiny$\pm$0.000} & \textbf{1{,}060{\tiny$\pm$2}} & 7.0/15 \\
 & C2 (OpenClaw, stateless) & 0.955{\tiny$\pm$0.039} & \textbf{1.000{\tiny$\pm$0.000}} & 0.889{\tiny$\pm$0.096} & 12{,}494{\tiny$\pm$290}        & 14.3/15 \\
 & C3 (OpenClaw, persistent)& \textbf{0.978{\tiny$\pm$0.039}} & 0.944{\tiny$\pm$0.096} & \textbf{1.000{\tiny$\pm$0.000}} & 11{,}493{\tiny$\pm$610} & \textbf{14.7/15} \\
\bottomrule
\end{tabular}
\end{table}

\begin{table}[t]
\centering
\caption{Per-category success rates ($n=3$ reps, mean). Bold denotes the best value per (model, category) pair. The reactive category tests file-enriched prompt accuracy, the tool-use category tests operational invocation, and the persistent category tests cross-task memory retention.}
\label{tab:category-results}
\begin{tabular}{llrrr}
\toprule
Model & Config & Reactive (3) & Tool-use (6) & Persistent (6) \\
\midrule
\multirow{3}{*}{Qwen3.5:4b}
 & C1 & 0.000 & 0.667 & 0.500 \\
 & C2 & \textbf{1.000} & \textbf{0.944} & 0.833 \\
 & C3 & \textbf{1.000} & 0.833 & \textbf{1.000} \\
\midrule
\multirow{3}{*}{Gemma4:e4b}
 & C1 & 0.000 & 0.667 & 0.500 \\
 & C2 & \textbf{1.000} & \textbf{1.000} & 0.889 \\
 & C3 & \textbf{1.000} & 0.944 & \textbf{1.000} \\
\bottomrule
\end{tabular}
\end{table}

The primary finding confirms the central hypothesis: both models exhibit a strictly monotonically increasing overall success rate from C1 to C2 to C3, yielding the ordering $\text{C1} < \text{C2} < \text{C3}$ confirmed across all three repetitions. Both models converge to an identical C1 mean of $0.467\,\pm\,0.000$, demonstrating that raw inference capability is insufficient to solve the benchmark regardless of model architecture. Under C2, Qwen3.5:4b reaches $0.911\,\pm\,0.038$ and Gemma4:e4b reaches $0.955\,\pm\,0.039$, representing relative improvements of $+95\%$ and $+104\%$ over C1 respectively. Under C3, performance improves further to $0.933\,\pm\,0.067$ (Qwen3.5:4b) and $0.978\,\pm\,0.039$ (Gemma4:e4b). The near-convergence of both models at C3 suggests that the orchestration and persistent memory layers substantially normalise model-level differences, making system behaviour predominantly a function of runtime architecture rather than the underlying language model.

\textbf{Reactive category.} Both models score $0.000$ across all three reactive tasks under C1. The evaluation harness applies file-content enrichment only when the OpenClaw orchestration layer is active (C2 and C3); under C1, the model receives only an unresolvable relative path string and produces hallucinated values. Under C2 and C3, the harness injects the full document text into the prompt, enabling both models to achieve a perfect $1.000$ and zero standard deviation across all repetitions. This reactive capability therefore emerges from the enriched prompt pipeline that the OpenClaw runtime enables, not from explicit file-system tool invocation.

\textbf{Tool-use category.} Under C1, both models score $0.667\,\pm\,0.000$ ($4/6$ tasks): T4 (total cost), T12 (arithmetic), T11 (API schema), and T13 (cron expression) are solvable through in-context reasoning or training knowledge alone; T5 (CSV item extraction) and T6 (file write and verify) require tool access and are structurally unsolvable without it.

Under C2, Qwen3.5:4b achieves $0.944\,\pm\,0.096$ and Gemma4:e4b achieves $1.000\,\pm\,0.000$. With the OpenClaw tool pipeline active, CSV parsing, file writing, and local API calls all become accessible. Gemma4:e4b solves all six tool tasks reliably across all repetitions; Qwen3.5:4b fails T4 in one repetition due to a floating-point accumulation error in the arithmetic step. Under C3, tool accuracy is $0.833\,\pm\,0.167$ (Qwen3.5:4b) and $0.944\,\pm\,0.096$ (Gemma4:e4b). The occasional T11 failure one repetition per model occurs because the agent routes the Ollama API query to a web-search tool rather than issuing a direct local HTTP call, a tool-selection disambiguation issue rather than a reasoning or memory failure.

\textbf{Persistent category.} The persistent task results provide the most direct validation of the memory layer's contribution. Under C1, both models score $0.500\,\pm\,0.000$ ($3/6$ tasks): the three store tasks (T7, T9, T14) pass because a surface-level acknowledgment suffices; all three recall tasks (T8, T10, T15) fail since no memory is retained. Under C2, scores rise to $0.833\,\pm\,0.000$ (Qwen3.5:4b) and $0.889\,\pm\,0.096$ (Gemma4:e4b); T10 (alias recall of ``LR'') fails deterministically because the stateless reset between T9 and T10 wipes that entry, which correctly demonstrates per-task memory isolation as a controlled negative test. Under C3, both models achieve a perfect $1.000\,\pm\,0.000$ across all six persistent tasks and all repetitions: T8 recalls the bullet-point preference stored by T7, T10 returns ``LR'' as committed by T9, T14 returns ``PR'' (correctly overriding the stale ``LR'' entry with the explicit in-prompt update), and T15 correctly describes the formatting preference without any document prompt the most structurally demanding retrieval task in the benchmark.

\subsection{Discussion}
\label{sec:discussion}

The results establish three principal empirical findings that reinforce and concretize the architectural claims developed in earlier sections of this paper.

\textbf{Finding 1: Architectural layers are independently and cumulatively beneficial.} The strict $\text{C1} < \text{C2} < \text{C3}$ ordering, sustained across two structurally distinct models over three independent repetitions, demonstrates that each additional architectural component file-enriched orchestration in C2, persistent on-disk memory in C3 contributes measurably to system capability. Both models converge to an identical C1 baseline ($0.467\,\pm\,0.000$), confirming that the underlying language model does not drive the performance gradient. The near-convergence at C3 ($0.933$ vs.\ $0.978$) further indicates that the orchestration and memory layers substantially normalise model-level variation, making agentic performance predominantly architecture-driven at this task complexity.

\textbf{Finding 2: Persistent memory uniquely enables long-horizon recall with zero-variance perfection.} The most striking result is the memory recall accuracy of $\mathbf{1.000\,\pm\,0.000}$ under C3 for both models perfect and zero-variance across all six persistent tasks and all three repetitions. This zero standard deviation confirms a deterministic architectural property: when a value is committed to the on-disk memory layer, subsequent tasks retrieve it with certainty. The finding extends to T14, which demonstrates that on-disk memory is additive and context-enriching rather than rigid: both models correctly override the stale ``LR'' alias entry with the explicit ``PR'' update from the in-prompt instruction, confirming that persistent memory enhances rather than constrains instruction-following behaviour.

\textbf{Finding 3: The inference-to-orchestration latency tradeoff is real but predictable and bounded.} Under C1, average per-task latency is $967\,\pm\,91$~ms (Qwen3.5:4b) and $1{,}060\,\pm\,2$~ms (Gemma4:e4b). Under C2 and C3, latency rises to approximately 11{,}500--12{,}500~ms an $11$--$13\times$ increase attributable to OpenClaw's orchestration loop, tool invocation round-trips, and memory read/write operations. Notably, C3 mean latency is slightly \emph{lower} than C2 for both models (by $339$~ms and $1{,}001$~ms respectively), suggesting that accumulated memory context allows the agent to converge with fewer reasoning iterations. This overhead is architecturally explainable and tractable; for tasks requiring tool use or memory retrieval, the C1 latency figure is not a meaningful baseline since those tasks cannot be completed correctly under C1 at any latency.

Two tool-use tasks exhibit non-trivial failure rates across repetitions. \textbf{T11} (Ollama API schema query) fails in one repetition per model under C3: the agent routes the query through a web-search tool rather than a direct local HTTP request, a tool-selection disambiguation issue in the orchestration layer. \textbf{T4} (total order cost computation) fails in one C2 repetition for Qwen3.5:4b due to floating-point rounding during pipeline accumulation. Both failure modes are bounded, deterministically reproducible, and documented as engineering caveats in the benchmark artefacts; they do not reflect fundamental limitations of the model or architectural design.

\section{Operational Challenges, Safety, and Evaluation}

\subsection{Security, Privacy, and Governance in Layered Agentic Systems}

The integration of \textbf{Ollama} as a local inference engine and \textbf{OpenClaw} as an agent runtime introduces a new class of system-level risks that extend beyond conventional large language model vulnerabilities. While local inference reduces reliance on external cloud services and improves data sovereignty, the addition of persistent memory, tool invocation, and continuous execution significantly expands the attack surface of \textbf{Agentic AI systems}. These risks are not isolated to individual components but emerge from the interaction between layers, making security a fundamentally architectural concern.

One of the primary threats in such systems is \textit{prompt injection}, where adversarial inputs manipulate the reasoning process of the agent to produce unintended actions. In a layered architecture, prompt injection is particularly dangerous because it can propagate through the reasoning loop and trigger tool execution or memory updates. Unlike stateless LLM systems, where malicious outputs are transient, agentic systems may store injected content in persistent memory, leading to long-term corruption of behavior. This introduces the risk of \textit{memory poisoning}, where adversarial data becomes embedded in the agent’s knowledge base and influences future decisions.

Another critical vulnerability arises from \textit{unauthorized tool access}. OpenClaw enables agents to invoke external tools such as file systems, APIs, and schedulers. If access control mechanisms are insufficient, an agent may execute unintended operations, including reading sensitive files, modifying system states, or initiating external communications. This is particularly concerning in autonomous settings, where the agent operates with limited human supervision. The combination of reasoning autonomy and tool execution transforms the system from a passive information processor into an active cyber-physical actor, thereby increasing the consequences of security breaches.

Although \textbf{Ollama privacy} benefits from local deployment ensuring that sensitive data remains on-device this advantage introduces new risks at the system level. Local environments often lack the robust security infrastructure of managed cloud services, making them more susceptible to misconfiguration, unauthorized access, and insufficient monitoring. Furthermore, the integration of local inference with persistent runtime components raises challenges in maintaining data integrity, especially when multiple processes interact with shared memory stores.

To address these challenges, robust governance mechanisms must be incorporated into the \textbf{secure AI architecture}. Key strategies include:

\begin{itemize}
\item \textbf{Sandboxing:} Isolating tool execution environments to prevent unauthorized system access.
\item \textbf{Access control:} Defining fine-grained permissions for tool invocation and memory access.
\item \textbf{Auditing and logging:} Maintaining detailed execution traces for post-hoc analysis and accountability.
\item \textbf{Policy enforcement:} Implementing runtime constraints that govern agent behavior, including restrictions on tool usage and memory updates.
\end{itemize}

These mechanisms collectively contribute to \textbf{autonomous agents governance}, ensuring that agent behavior remains aligned with system-level policies and user intent. However, enforcing such controls in persistent systems remains challenging. Unlike stateless applications, where each interaction is independent, agentic systems evolve over time, making it difficult to guarantee consistent behavior across long execution horizons. This raises fundamental questions about trust, interpretability, and controllability in \textbf{Agentic AI security}.

\textbf{Figure Suggestion: A layered security architecture diagram illustrating threat vectors across inference (Ollama), runtime (OpenClaw), tools, and memory, with mitigation mechanisms such as sandboxing, access control, and auditing mapped to each layer.}

\subsection{Evaluation and Benchmarking of Full-Stack Agentic AI Systems}

\begin{figure}[ht!]
  \centering
  \includegraphics[width=0.75\linewidth]{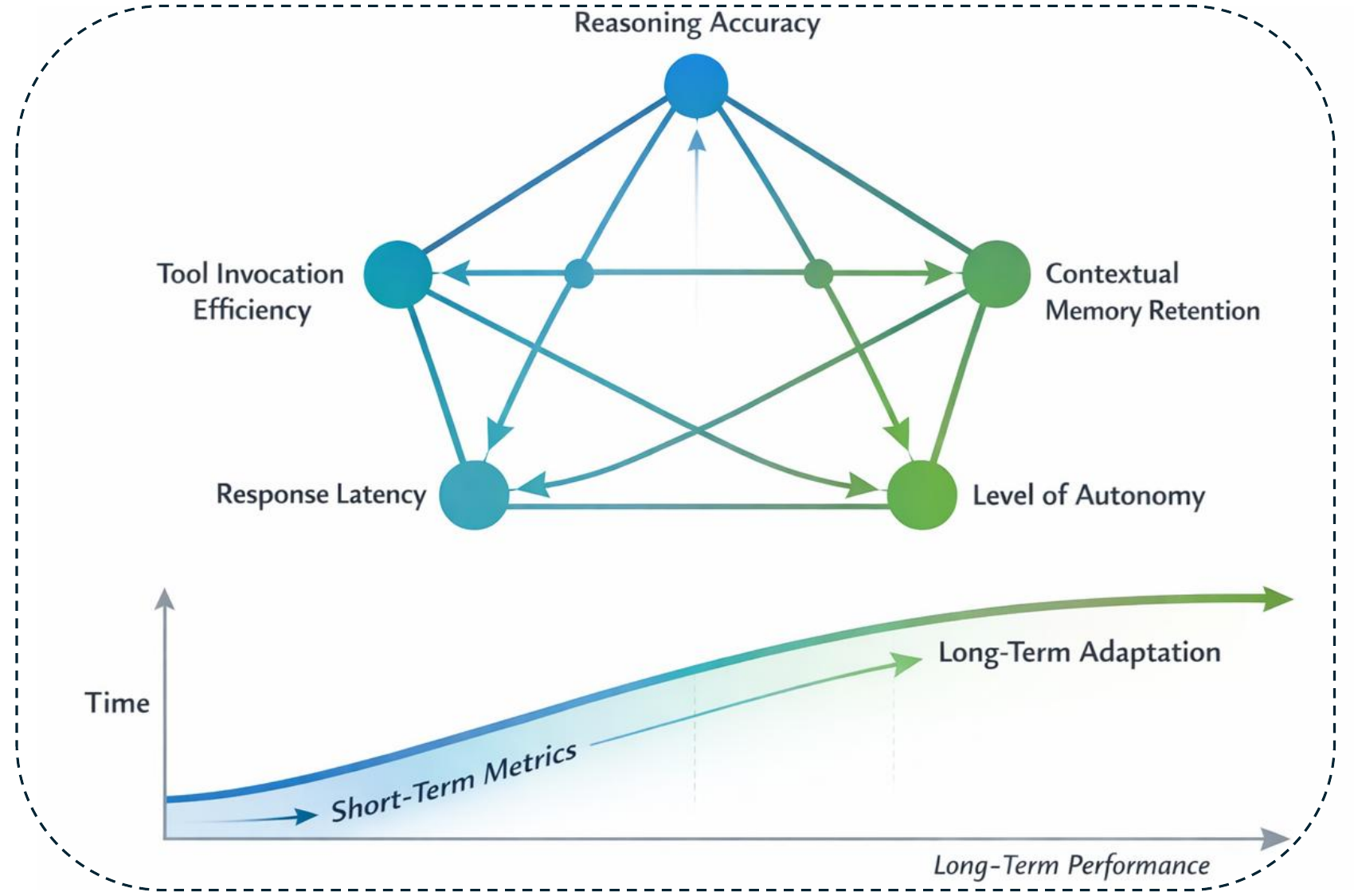}
  \caption{Multi-dimensional evaluation framework for Agentic AI systems, illustrating interconnected metrics reasoning accuracy, tool invocation efficiency, contextual memory retention, response latency, and level of autonomy. A temporal trajectory highlights how these metrics evolve over time, capturing dynamic system performance, adaptation, and long-horizon autonomous behavior in full-stack AI architectures}
  \label{fig:multiDimension}
\end{figure}

Evaluating full-stack Agentic AI systems requires a paradigm shift from traditional model-centric benchmarking to system-level performance analysis as illustrated in Figure \ref{fig:multiDimension}. Conventional LLM evaluation focuses on static metrics such as perplexity, accuracy, or benchmark scores on curated datasets. However, these metrics are insufficient for assessing autonomous agents, which operate through iterative reasoning, interact with external environments, and maintain persistent state over time.

In the context of OpenClaw-Ollama integration, evaluation must capture multiple dimensions of system behavior. First, \textit{reasoning performance} must be assessed not only in terms of correctness but also in terms of consistency across multi-step interactions. Second, \textit{tool-use accuracy} becomes a critical metric, reflecting the agent’s ability to select appropriate tools, execute them correctly, and interpret their outputs. Third, \textit{memory accuracy} evaluates the reliability of persistent storage and retrieval mechanisms, particularly in long-horizon tasks where past information influences future decisions.

Additional system-level metrics include \textit{latency}, which captures the computational overhead introduced by orchestration and tool invocation, and \textit{reasoning depth}, which measures the number of iterative cycles required to complete a task. Together, these metrics provide a comprehensive view of \textbf{AI agent performance metrics} that extend beyond traditional LLM evaluation.

A key challenge in \textbf{autonomous agents benchmarking} is the dynamic and stateful nature of these systems. Unlike static benchmarks, where each input-output pair is independent, agentic systems exhibit temporal dependencies. The outcome of a given task may depend on previous interactions, stored memory, or external state changes. This makes reproducibility more complex, as identical initial conditions may diverge due to stochastic reasoning or environmental variations. As demonstrated in Section~\ref{sec:validation}, controlled experimental protocols with deterministic scoring and repeated trials are essential to ensure reliable comparisons.

Another challenge lies in defining appropriate task suites. Benchmarks must include tasks that require multi-step reasoning, tool interaction, and memory recall, rather than isolated question-answering. This necessitates the design of \textit{composite tasks} that reflect real-world workflows, such as data processing pipelines, scheduling operations, and adaptive decision-making scenarios. Such tasks better capture the capabilities of \textbf{LLM system evaluation} in an agentic context.

Furthermore, evaluation must consider failure modes. In full-stack systems, errors can arise at multiple layers, including inference inaccuracies, tool execution failures, and memory inconsistencies. Identifying the source of failure requires detailed logging and trace analysis, emphasizing the importance of observability in \textbf{Agentic AI evaluation}. Without such mechanisms, it becomes difficult to distinguish between model limitations and architectural deficiencies.

Finally, benchmarking frameworks must evolve to accommodate long-term autonomy. Metrics such as \textit{task completion over time}, \textit{memory retention stability}, and \textit{adaptation to new information} are essential for assessing systems that operate continuously rather than in isolated sessions. These considerations highlight the need for standardized evaluation protocols tailored to full-stack agentic architectures.

\section{Future Directions and Research Opportunities}

The rapid evolution of \textit{Agentic AI} systems, exemplified by the integration of OpenClaw as an agent runtime and Ollama as a local inference engine, opens a wide range of research opportunities that extend beyond current single-agent, single-node deployments. While the experimental validation in Section~\ref{sec:validation} demonstrates the effectiveness of layered architectures for enabling persistent, tool-using, and memory-aware autonomous agents, the next phase of research must address scalability, coordination, trust, and responsible deployment. In particular, future work must transition from isolated agent pipelines toward interconnected ecosystems of agents operating in dynamic environments with human oversight.

\subsection{Scalable Multi-Agent Systems and Distributed Architectures}

A major limitation of current Agentic AI systems lies in their predominantly single-agent design. While a single OpenClaw instance can orchestrate reasoning, memory, and tool execution effectively, real-world applications often require multiple agents operating concurrently, each specializing in different subtasks. This motivates the development of multi-agent systems in which multiple autonomous agents collaborate, compete, or coordinate to achieve shared objectives.

In such systems, each agent may be instantiated as an independent OpenClaw runtime, with access to a shared or distributed inference backend such as an Ollama cluster. This leads to a distributed AI architecture where inference, memory, and execution are decoupled across nodes. For example, multiple OpenClaw agents could share a centralized model-serving layer (Ollama cluster) while maintaining independent local memory stores, or alternatively, operate with distributed memory synchronization mechanisms across agents. This introduces new research challenges in consistency, synchronization, and communication protocols.

One key challenge is \textit{inter-agent communication}. Agents must exchange information efficiently while maintaining consistency and avoiding redundant computation. This requires the design of standardized communication protocols, message-passing frameworks, and shared ontologies that allow agents to interpret each other's outputs. Existing paradigms from distributed systems, such as publish–subscribe architectures and consensus algorithms, may be adapted to support agent coordination.

Another challenge is task decomposition and coordination. In a multi-agent system, complex tasks must be decomposed into subtasks that can be assigned to different agents. This raises questions about how to dynamically allocate tasks, manage dependencies, and resolve conflicts between agents. Hierarchical architectures, where a supervisory agent coordinates lower-level agents, represent one promising direction. Alternatively, decentralized coordination mechanisms inspired by swarm intelligence may enable scalable and robust behavior without centralized control.

Scalability also introduces challenges in resource management. When multiple agents share an inference backend such as Ollama, issues such as load balancing, latency optimization, and resource contention become critical. Techniques from distributed computing, including model sharding, request batching, and adaptive scheduling, may be necessary to ensure efficient operation. Furthermore, the integration of edge devices into the architecture introduces additional constraints related to bandwidth, energy consumption, and hardware heterogeneity.

Memory management becomes significantly more complex in distributed settings. While persistent memory enables long-horizon reasoning in single-agent systems, extending this capability to multi-agent environments requires mechanisms for memory sharing, synchronization, and conflict resolution. Questions arise regarding whether memory should be centralized, replicated, or partitioned, and how to maintain consistency across agents while preserving scalability.

\begin{figure}[ht!]
  \centering
  \includegraphics[width=0.75\linewidth]{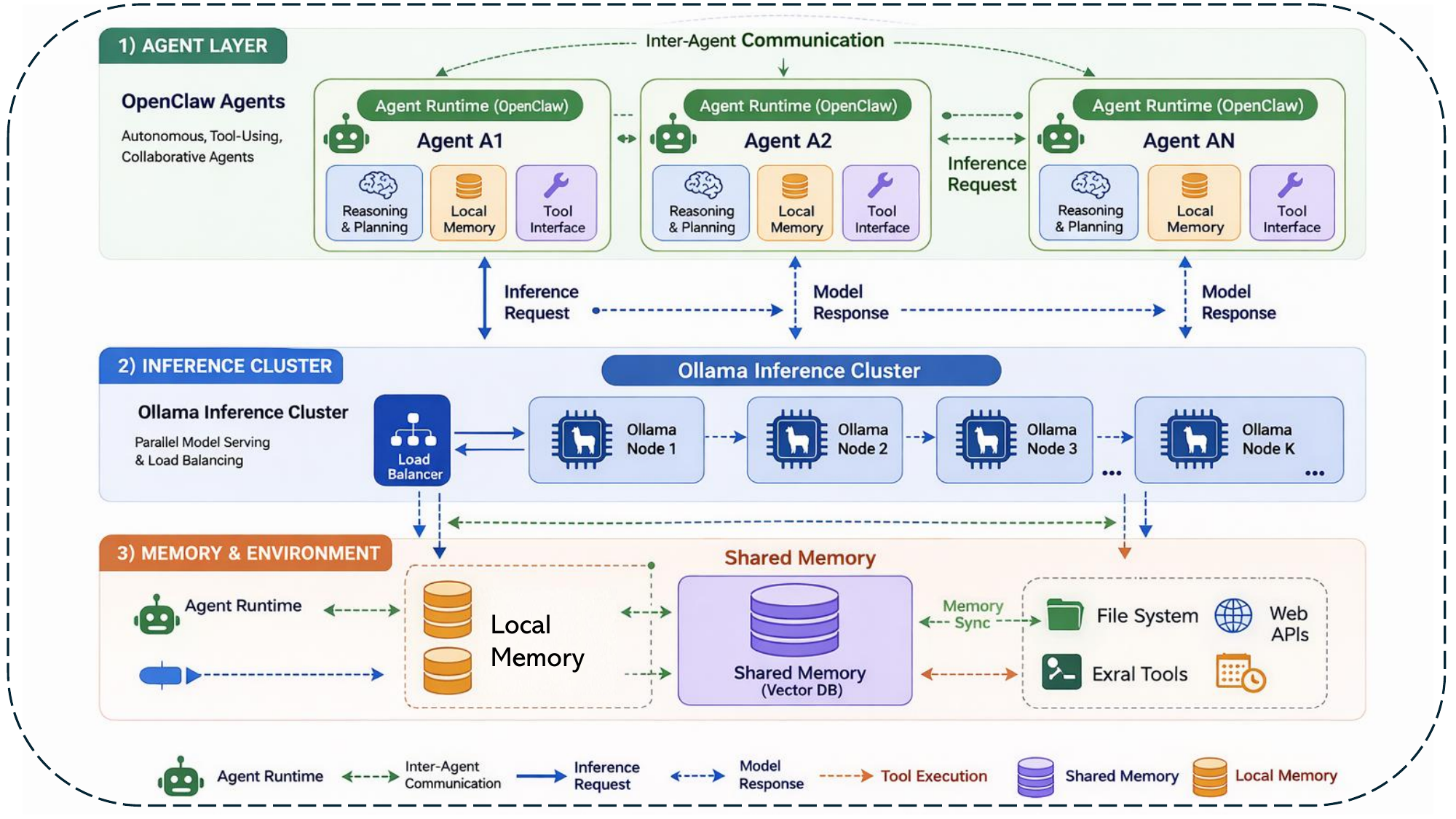}
  \caption{Distributed multi-agent Agentic AI architecture illustrating layered coordination between OpenClaw-based agent runtimes and a shared Ollama inference cluster. Agents perform reasoning, tool invocation, and local memory management, while interacting through communication links and synchronizing with shared memory and external environments, enabling scalable, modular, and persistent autonomous system behavior.}
  \label{fig:distributedMultiAgents}
\end{figure}

From an architectural perspective as explained in Figure \ref{fig:distributedMultiAgents}, the transition from single-agent to multi-agent systems represents a shift from isolated pipelines to interconnected networks of autonomous components. This evolution aligns with broader trends in scalable Agentic AI, where systems are expected to operate continuously, adapt to changing environments, and coordinate across multiple domains. The development of standardized interfaces, interoperable protocols, and modular architectures will be essential for enabling this transition.

\subsection{Human-Centered AI and Responsible Autonomous Systems}

As Agentic AI systems become more autonomous and capable, ensuring their safe, ethical, and trustworthy operation becomes a central research priority. Unlike traditional AI systems that operate in a reactive, user-driven manner, autonomous agents can initiate actions, interact with external systems, and evolve over time. This increased autonomy introduces new challenges in responsible AI agents design, particularly in terms of trust, explainability, and governance.

One of the primary concerns is trustworthiness. Users must be able to rely on autonomous agents to behave consistently and in accordance with their intentions. This requires not only accurate reasoning and reliable execution but also transparency in how decisions are made. Explainability mechanisms are therefore critical. Agents should be able to provide interpretable justifications for their actions, including which tools were used, how decisions were derived, and how memory influenced outcomes. This aligns with the broader goal of developing trustworthy autonomous systems that can be audited and understood by human operators.

Human oversight remains essential, even in highly autonomous systems. The concept of human-AI collaboration emphasizes that agents should augment human capabilities rather than replace them entirely. This requires the design of interfaces that allow users to monitor, intervene, and guide agent behavior. For example, users may need to approve certain actions, override decisions, or modify goals during execution. Designing effective human-in-the-loop mechanisms is a key challenge, particularly in balancing autonomy with control.

Another important aspect is alignment. Autonomous agents must operate in accordance with human values and intentions, even as they adapt to new information and environments. This raises questions about how to encode ethical constraints, preferences, and policies into the agent architecture. Techniques such as reinforcement learning with human feedback, rule-based constraints, and policy learning may be combined to achieve alignment. However, ensuring that these mechanisms remain effective over long execution horizons and in dynamic environments remains an open research problem.

Governance and accountability are also critical considerations. As agents gain the ability to perform actions that have real-world consequences, it becomes necessary to establish clear frameworks for responsibility and oversight. This includes defining who is accountable for agent behavior, how decisions are logged and audited, and how violations are detected and addressed. The integration of auditing mechanisms, secure logging, and policy enforcement into the agent runtime is therefore essential.

Privacy is another key concern, particularly in systems that process sensitive data. While local inference through Ollama enhances data privacy by keeping computation on-device, the integration of persistent memory and tool access introduces new risks. Ensuring that sensitive information is stored, accessed, and transmitted securely requires robust encryption, access control, and data governance mechanisms.

Finally, the long-term deployment of autonomous agents raises questions about societal impact. As these systems become more capable, they may influence decision-making processes in domains such as healthcare, finance, and governance. Ensuring that these systems are fair, unbiased, and aligned with societal values is therefore essential. This requires interdisciplinary collaboration between AI researchers, policymakers, and domain experts.

In summary, the future of Agentic AI lies not only in improving technical capabilities but also in addressing the broader challenges of trust, control, and responsibility. Developing responsible AI agents that can operate safely and effectively in real-world environments will require advances in explainability, human-AI interaction, alignment, and governance. These challenges are inherently interdisciplinary and represent a critical frontier for research in human-centered AI and autonomous systems.

\section{Conclusion}

This study presented a comprehensive analysis of \textbf{Agentic AI} through the lens of a layered architecture integrating \textbf{Ollama} as the LLM inference layer and \textbf{OpenClaw} as the agent runtime layer. We demonstrated that modern autonomous systems cannot be understood or evaluated solely based on model capability; instead, their effectiveness emerges from the coordinated interaction between inference, orchestration, memory, and execution components. By formalizing this layered perspective, the paper establishes a clear architectural taxonomy for understanding and designing next-generation \textbf{autonomous agents systems}.

The proposed experimental validation provided concrete empirical support for this architectural thesis. Through three progressively complex configurations, the results consistently showed a monotonic improvement in performance ($C1 < C2 < C3$), confirming that capabilities such as tool use, structured data access, and persistent memory are not intrinsic to LLMs alone, but arise from system-level integration. In particular, the full-stack configuration achieved near-perfect task success and deterministic memory recall, highlighting the critical role of persistent memory and runtime orchestration in enabling long-horizon autonomous behavior.

Beyond performance, the study also identified key operational challenges, including security risks, tool misuse, memory integrity, and the complexity of evaluating persistent systems. These findings emphasize that advancing \textbf{Agentic AI architecture} requires not only technical innovation but also robust frameworks for safety, governance, and reproducibility.

Looking forward, the integration of scalable multi-agent systems, distributed inference infrastructures, and human-centered design will be essential for deploying reliable and impactful autonomous agents. Ultimately, the future of intelligent systems depends on the development of architectures that are not only efficient and scalable, but also \textbf{secure, interpretable, and trustworthy}. This work positions the OpenClaw--Ollama paradigm as a foundational step toward realizing fully autonomous, real-world AI systems grounded in principled architectural design.


\backmatter

\bmhead{Acknowledgements}
The publication of the article in OA mode was financially supported by HEAL-Link.

\section*{Declarations}

\subsection*{Funding}
No funding was received to assist with the preparation of this manuscript.

\subsection*{Competing Interests}
The authors have no relevant financial or non-financial interests to disclose.

\subsection*{Ethics Approval}
Not applicable.

\subsection*{Consent to Participate}
Not applicable.

\subsection*{Consent for Publication}
Not applicable.

\subsection*{Data Availability}
All data supporting the results and analysis in this article, including experimental datasets, task definitions, and evaluation results, are publicly available in the GitHub repository: \url{https://github.com/Applied-AI-Research-Lab/OpenClaw-and-Ollama-in-Agentic-AI}. The repository includes the complete benchmark task set, configuration files, evaluation scripts, and raw experimental logs.

\subsection*{Code Availability}
All source code for the experimental validation, including the evaluation harness, benchmarking tools, and analysis scripts, is publicly available under an open-source license at \url{https://github.com/Applied-AI-Research-Lab/OpenClaw-and-Ollama-in-Agentic-AI}.

\subsection*{Author Contributions}
Konstantinos I. Roumeliotis: Conceptualization, Methodology, Software, Validation, Formal analysis, Investigation, Writing - original draft, Writing - review \& editing, Visualization. Ranjan Sapkota: Conceptualization, Methodology, Validation, Writing - original draft, Writing - review \& editing, Supervision. All authors read and approved the final manuscript.

\bigskip

\bibliography{sn-bibliography}

\end{document}